\documentclass[journal]{IEEEtran}
\usepackage{graphicx}
\usepackage{tabularx}
\usepackage{cite}
\usepackage{amsmath}
\usepackage{amssymb}
\usepackage{textcomp}
\usepackage{enumitem}
\usepackage{booktabs}
\usepackage{multirow}
\usepackage{siunitx}
\usepackage{balance}
\usepackage{xcolor}
\usepackage[para,online,flushleft]{threeparttable}
\usepackage[linesnumbered, noend, algoruled,boxed,lined]{algorithm2e}

\DeclareMathOperator\supp{supp}
\DeclareMathOperator\inff{inf}
\DeclareMathOperator\sgn{sgn}
\DeclareMathOperator\iif{if}

\ifCLASSINFOpdf
\else
\fi

\begin{document}

\title{Hard-ODT: Hardware-Friendly Online Decision Tree Learning Algorithm and System}

\author{Zhe~Lin,~\IEEEmembership{Member,~IEEE,}
        Sharad~Sinha,~\IEEEmembership{Member,~IEEE,}
        and~Wei~Zhang,~\IEEEmembership{Member,~IEEE}\vspace{-8mm}
\thanks{The author Zhe Lin (linzh01@pcl.ac.cn; zlinaf@connect.ust.hk) is with the AI Research Center, Peng Cheng Laboratory, Shenzhen, China. This work was done while he was in Reconfigurable Computing Systems Lab (RCSL) at Hong Kong University of Science and Technology (HKUST).}
\thanks{The author Sharad Sinha (sharad\_sinha@ieee.org) is with the Dept. of Computer Science and Engineering, Indian Institute of Technology (IIT) Goa.}
\thanks{The author Wei Zhang (wei.zhang@ust.hk) is with RCSL, the Dept. of Electronic and Computer Engineering, HKUST.}}
\maketitle

\begin{abstract}
Decision trees are machine learning models commonly used in various application scenarios. In the era of big data, traditional decision tree induction algorithms are not suitable for learning large-scale datasets due to their stringent data storage requirement. Online decision tree learning algorithms have been devised to tackle this problem by concurrently training with incoming samples and providing inference results. However, even the most up-to-date online tree learning algorithms still suffer from either high memory usage or high computational intensity with dependency and long latency, making them challenging to implement in hardware. To overcome these difficulties, we introduce a new quantile-based algorithm to improve the induction of the Hoeffding tree, one of the state-of-the-art online learning models. The proposed algorithm is light-weight in terms of both memory and computational demand, while still maintaining high generalization ability. A series of optimization techniques dedicated to the proposed algorithm have been investigated from the hardware perspective, including coarse-grained and fine-grained parallelism, dynamic and memory-based resource sharing, pipelining with data forwarding. Following this, we present Hard-ODT, a high-performance, hardware-efficient and scalable online decision tree learning system on a field-programmable gate array (FPGA) with system-level optimization techniques. Performance and resource utilization are modeled for the complete learning system for early and fast analysis of the trade-off between various design metrics. Finally, we propose a design flow in which the proposed learning system is applied to FPGA run-time power monitoring as a case study. Experimental results show that our proposed algorithm outperforms the state-of-the-art Hoeffding tree learning method, leading to 0.05\% to 12.3\% improvement in inference accuracy. Real implementation of the complete learning system on the FPGA demonstrates a 384$\times$ to 1581$\times$ speedup in execution time over the state-of-the-art design. The power modeling strategy with Hard-ODT achieves an average power prediction error within 4.93\% of a commercial gate-level power estimation tool.
\end{abstract}

\begin{IEEEkeywords}
FPGA, online learning, decision tree, quantile, power modeling.
\end{IEEEkeywords}

\IEEEpeerreviewmaketitle

\vspace{-3mm}
\section{Introduction}
Decision tree algorithms are a popular class of machine learning algorithm and have been deployed in many real scenarios~\cite{cvprchen16, kaneko18, oliv18}, especially when multiple decision trees are combined into powerful ensemble models, such as XGBoost~\cite{xgboost} and random forests~\cite{brei01}. Recently, the ensemble of decision trees as deep forests~\cite{zhou17} has been reported to produce comparable performance compared to deep neural networks. However, there are several drawbacks that limit the full exploitation of the traditional decision trees (e.g., IDT3~\cite{idt3}, CART~\cite{cart} and C4.5~\cite{quin14}). The first drawback is the extensive memory consumption during the training process, which is proportional to the size of datasets. Classic decision tree learners assume that the complete datasets can be preloaded before training starts. This reduces their capability to train with large-scale datasets, especially when, nowadays, large amount of data is being generated daily. The second disadvantage comes with the learners' inability to adapt themselves to new data once the training process is terminated. In the era of big data, the size of datasets is no longer the bottleneck of learning algorithms. Instead, the ability to effectively learn from massive data and rationally make use of incoming data becomes more fundamental and critical.

To broaden the applicability of decision tree algorithms, extensions from traditional tree algorithms to batch learning and online learning (or so-called incremental learning) have been studied, which aim at adapting the models to incoming data without losing previously learned knowledge. One of the state-of-the-art online learning methods for streaming data is the \textit{Hoeffding tree}~\cite{pedro00} algorithm and its variants~\cite{vfdtc, vfml, streamdm, ga08, hul01, ikno11, kour16, vasi17}. The Hoeffding tree presents an enhancement of the decision tree induction algorithm which leverages the accumulated samples to estimate the complete datasets statistically. It is capable of performing training and inference concurrently. The Hoeffding tree is widely used in various application scenarios~\cite{bar14, fai15, wu16, nie17}.

While efficient software implementation has been investigated for processors to accelerate the Hoeffding tree~\cite{vfml, streamdm}, there are still many hindrances to the compact implementation and optimization of the Hoeffding tree design from the hardware perspective. We identify two principal challenges limiting Hoeffding tree implementation in hardware: 1) the high cost of memory usage to store the required subset of samples as well as characteristics in each leaf node; and 2) the high computational demand with dependency and long latency between iterations in the learning process, which can hamper efficient data processing with optimization schemes such as parallelism and pipelining. Furthermore, we observe a trade-off between the above two factors in the state-of-the-art designs: the methods in~\cite{streamdm} and~\cite{ga08}, attempting to reduce the memory usage, tend to extensively increase the computational intensity and latency, and vice versa, as in the proposed methods of~\cite{vfdtc} and~\cite{vfml}. The high and unbalanced need of memory and computation makes the existing approaches difficult to efficiently implement in hardware, especially on FPGAs where memory and digital signal processing (DSP) resources are both limited. Motivated by the above challenges and observations, we seek opportunities to implement and optimize the Hoeffding tree in a hardware-friendly and scalable way, and also strive to make use of resources in a more balanced manner. In this paper, we propose Hard-ODT, the first implementation of the Hoeffding tree learning system on FPGA, with the following contributions:
\begin{itemize}
\item We first introduce a quantile-based algorithm for Hoeffding tree induction, which uses light-weight computation and constant memory, while preserving high accuracy.
\item We present hardware optimization techniques dedicated to the proposed algorithm, in order to achieve high hardware efficiency and scalability. These includes different levels of parallelism, dynamic and memory-based resource sharing, and pipelining with data forwarding.
\item We investigate optimization techniques for tree growing, categorical attribute learning and split judgment to establish the complete online decision tree system on FPGA.
\item We model performance and resource utilization for the proposed online decision tree learning system on FPGA for fast evaluation of the critical design metrics.
\item We develop a design flow to apply the proposed online learning system to FPGA run-time power monitoring.
\end{itemize}

\section{Algorithm and Challenges}
\setlength{\abovedisplayskip}{3pt}
\setlength{\belowdisplayskip}{3pt}
\subsection{Hoeffding Tree Induction Algorithm}
\begin{figure}[t]
	\begin{center}
		\includegraphics[width=0.8\linewidth]{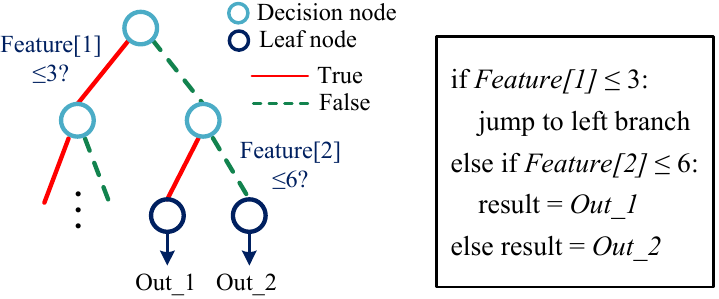}
		\vspace{-2mm}
		\caption{Graphical and textual representation of a decision tree.}
		\label{fig: dtback}
		\vspace{-5mm}
	\end{center}
\end{figure}

The decision tree~\cite{rok08} learns the samples in the form of a tree structure. A tree node can be categorized as (1) a leaf/terminal node --- a node associated with an output result (i.e., a class label for classification or a value for regression) --- or (2) a decision/internal node --- an intermediate node to decide on one of its child nodes to go to. The training process is to determine an if-then-else decision rule for every decision node and an output value for every leaf node, simply based on a certain split criterion computed with all the samples gathered in the corresponding nodes. To make a new inference for an unsolved case, the decision tree firstly starts with the root node and moves each sample to child nodes iteratively until a leaf node with inference result is reached, as shown in Fig.~\ref{fig: dtback}. The induction flows of the online and offline decision tree algorithms differ in the ways they make the split decisions: the offline decision trees make split decisions with well-defined datasets, while the online decision trees make the decisions dynamically with an incoming data stream. 

As one of the state-of-the-art online decision tree algorithms, the Hoeffding tree exploits the potential for the currently seen sample set to represent an infinite sample set when making split decisions, as described in Algorithm~\ref{alg: ht}. At each leaf node, the \textit{Hoeffding bound} (a.k.a. Chernoff bound)~\cite{hoeffb} is used to tell how close the current best split approaches the optimal split given an infinite sample set. Suppose we make $n$ independent observations of a random variable $ r$ within range $R$. The Hoeffding bound guarantees that the true mean $\overline{r}$ of $r$ will be at least $E[r]-\epsilon$, with
\begin{equation}
\footnotesize
\label{eq: hb}
\epsilon = \sqrt{\frac{R^2\ln(1/\delta)}{2n}}.
\end{equation}

Let $G(a_i)$ be the best measurement (e.g., gini impurity reduction) of a chosen split attribute $a_i$. The Hoeffding tree searches for the best and second-best $G(\cdot)$ values amongst all attributes. Given the sample set of size $n$ for a specific node and a desired $\delta$, the Hoeffding bound justifies that the current best attribute is the exact best attribute from an infinite dataset with probability $1-\delta$, if the following equation is satisfied:
\begin{equation}
\footnotesize
\label{eq: htb}
G(Best\ attr.)-G(2^{nd}\ Best\ attr.)>\sqrt{\frac{R^2\ln(1/\delta)}{2n}}.
\end{equation}

An additional tie condition is applied: when the two best attributes have close $G(\cdot)$, a split is taken if the Hoeffding bound is lower than a certain threshold $\tau$. That is,
\begin{equation}
\footnotesize
\label{eq: htbtie}
G(Best\ attr.)-G(2^{nd}\ Best\ attr.) < \sqrt{\frac{R^2\ln(1/\delta)}{2n}} < \tau.
\end{equation}

\vspace{-9mm}
\subsection{Challenges}
\label{subsec: chalg}
\LinesNumbered
\begin{algorithm}[t]
	\footnotesize
	\caption{Traditional Hoeffding tree algorithm}
	\label{alg: ht}
	\SetKwInOut{Input}{input}
	\SetKwInOut{Output}{output}
	\Input{samples denoted as $(x,y)$}
	\Output{Hoeffding tree denoted as $HT$}
	\For{each $(x_t, y_t)$ coming at time $t$}{
		filter $(x_t,y_t)$ to leaf $l$ of $HT$\\
		sample number in leaf $l$: $n_l \leftarrow n_l+1$\\
		update bin count $(attr_i, val_j, class_k)$ $n_{ijk}$ in leaf $l$\\
		\If{split trial is activated}{
			compute left/right partitions according to $n_{ijk}$\\
			compute $G(\cdot)$ for each attribute\\
			\If{$G(best)$\ -\ $G(2^{nd}\ best) > \sqrt{\frac{R^2\ln(1/\delta)}{2n_l}}$ or $\sqrt{\frac{R^2\ln(1/\delta)}{2n_l}} < \tau $}{
				Split leaf $l$ on the best attr.\\
				Initialize count $n_{ijk}$ for each leaf\\
			}
		}
	}
\end{algorithm}

\setlength{\textfloatsep}{0.5mm}
\setlength{\floatsep}{0.5mm}

\LinesNumbered
\begin{algorithm}[t]
	\footnotesize
	\caption{Incremental Gaussian approximation}
	\label{alg: gausapp}
	\SetKwInOut{Input}{input}
	\SetKwInOut{Output}{output}
	\Input{samples denoted as $(attr_{val}, weight)$}
	\Output{mean of Gaussian approximation denoted as $M$}
	\Output{variance of Gaussian approximation denoted as $V$}
	weight sum: $w\_sum \leftarrow first\ weight$ \\
	variance sum: $v\_sum \leftarrow 0$ \\
	$M \leftarrow first\ attr_{val}$ \\
	\For{each sample $(attr_{val}, weight)$ in sample set}{
		$w\_sum \leftarrow w\_sum + weight$ \\
		$M_{prior} \leftarrow M$ \\
		$M \leftarrow M + \frac{attr_{val} - M_{prior}}{w\_sum}$ \\
		$v\_sum \leftarrow v\_sum + (attr_{val} - M_{prior}) \times (attr_{val} - M)$ \\
		$V \leftarrow \frac{v\_sum}{w\_sum-1}$ \\
	}
\end{algorithm}

Studies~\cite{vfdtc, vfml, streamdm, ga08} have introduced several methods to improve the basic Hoeffding tree algorithm. These methods, however, reveal two main challenges for hardware implementation.

\textbf{1. High Cost of Memory Utilization.} In VFML~\cite{vfml}, both numeric and categorical attribute values are preserved in a fixed number of bins (denoted as $n_{ijk}$) in a first-come-first-served manner. If all the bins are occupied, the newly coming attribute values unseen in all the bins are simply discarded during runtime. Although this method works well with categorical attributes of which values are discrete and the total number can be determined in the compile time, it requires a bin of large size to fit each numeric attribute per class per node to achieve a wide value coverage. Hence, the memory requirement grows significantly with the number of attributes. This similarly exists in the method~\cite{ga08} using Greenwald and Khanna summaries~\cite{green01}, which requires to construct sample distribution from up to thousands of tuples per attribute-class combination per node. The exhaustive binary tree method~\cite{vfdtc} also suffers from injudicious use of memory because it needs to dynamically allocate memory for sample storage. 

\textbf{2. High Computational Intensity with Dependency and Long Latency.} To reduce memory utilization, Gaussian-based methods~\cite{ga08, streamdm} are applied to trade much higher computational intensity for memory efficiency. For each numeric attribute per class, the sample distribution is estimated in a form of Gaussian distribution. As the Gaussian function is determined by only two values, namely, mean and variance, the memory usage can be significantly compressed to $\# attribute \times \# class \times 2$ per node. However, the incremental update process of the mean and variance leads to high computational demand, as shown in Algorithm~\ref{alg: gausapp}. The requirement of computation resources is proportional to both the number of attributes and classes. Besides this, the split judgment stage also requires computing the cumulative density functions (CDFs) at each split point, which entails even higher computational power. Moreover, the update process incurs long latency and should be in order of time if the two successive iterations work on the same label. In addition to the high computational intensity, the long latency and data dependency further hinder this method from being effectively optimized in hardware.

\vspace{-3mm}
\section{Methodology}
As BRAM and DSP are limited resources for FPGAs, the excessive use of either on-chip memory or computation units in the aforementioned methods~\cite{vfml, vfdtc, ga08, streamdm} is neither efficient nor scalable while handling numeric attributes. The two design challenges described above and their interplay should be taken into consideration for joint optimization. To this end, we propose to introduce an up-to-date quantile algorithm in the induction of online decision trees. 

\vspace{-3mm}
\subsection{Quantile Estimation Using Asymmetric Signum Functions}
\LinesNotNumbered
\begin{algorithm}[t]
	\footnotesize
	\caption{Hoeffding tree induction with quantiles}
	\label{alg: qtht}
	\SetKwInOut{Input}{input}
	\SetKwInOut{Output}{output}
	\Input{streaming samples denoted as $(x, y)$}
	\Output{Hoeffding tree structure denoted as $HT$}
	Let $a_i\ (1 \leq i \leq |A|)$ denote the attribute in set A \\
	Let $c_j\ (1 \leq j \leq |C|)$ denote the class in set C \\
	Let ${\alpha}_k (1 \leq k \leq |Q|)$ denote the quantile index \\
	\setcounter{AlgoLine}{0}
	\nl \For{each $(x_t, y_t)$ $\in$ sample set}{
		\nl filter $(x_t, y_t)$ to leaf $f$ of $HT$ \\
		\nl sample num. at $f$: $n_f \leftarrow n_f+1$ \\
		\nl \For{j from 1 to $|C|$}{
			\nl  sample num. in class $j$: $n_{fj} \leftarrow (y_t ==j)\ ?\ n_{fj}+1 : n_{fj}$ \\
		}
		\nl \For{i from 1 to $|A|$}{
			\nl max. attr. value: $\max_{a_i} \leftarrow (a_i > \max_{a_i}) \ ? \ a_i\ :\ \max_{a_i}$ \\
			\nl min. attr. value: $\min_{a_i} \leftarrow (a_i < \min_{a_i}) \ ? \ a_i\ :\ \min_{a_i}$ \\
			\nl \For{j from 1 to $|C|$}{
				\nl \If{$y_t == j$}{
					\nl \For{k from 1 to $|Q|$}{
						\nl ${Q}_{ijt}({\alpha}_k) \leftarrow {Q}_{ij{t-1}}({\alpha}_k) - \lambda \sgn_{\alpha}({Q}_{ij{t-1}}({\alpha}_k)-a_i)$ \\
					}
				}
			}
		}
		\nl \If{split trial is activated}{
			\nl \For{i from 1 to $|A|$}{
				\nl \For{p from 1 to $|P|$}{
					\nl $pt \leftarrow \frac{\max_{a_i}-\min_{a_i}}{|P|+1}\times p + \min_{a_i}$ \\
					\nl \For{j from 1 to $|C|$}{
						\nl left distribution $L$: $dist_{Lij}(pt) \leftarrow 0$ \\
						\nl \For{k from 1 to $|Q|$}{
							\nl $dist_{Lij}(pt) \leftarrow (pt > {Q}_{ijt}({\alpha}_k))\  ?\ dist_{Lij}(pt) + 1 : dist_{Lij}(pt)$ \\
						}
						\nl  $dist_{Lij}(pt) \leftarrow \frac{dist_{Lij}(pt)}{|P|} \times n_{fj} $\\
						\nl $dist_{Rij}(pt) \leftarrow n_{fj} - dist_{Lij}(pt)$ \\
					}
				}
				\nl compute $G(a_i)$ for all $pt$ \\
			}
			\nl\If{$G(best)$-$G(2^{nd}\ best) > \sqrt{\frac{R^2\ln(1/\delta)}{2n_f}}$ or $\sqrt{\frac{R^2\ln(1/\delta)}{2n_f}} < \tau $}{
				\nl split $l$ on the best attr \& initialize new leaves\\
			}
		}
	}
\end{algorithm}

Quantiles~\cite{rob96} are cutting points dividing the range of a probability distribution into a certain number of intervals with equal probabilities. The quantile function $Q(\cdot)$ of a continuous variable is defined as the inverse of the CDF, $F(z) = Pr(x_t \leq z)$. Specifically, $Q(\cdot)$ can be written as
\begin{equation}
\footnotesize
\label{eq:qt}
Q(\alpha)=F_X^{-1}(\alpha)=\inff\{x\in \supp(F_X):\alpha\leq F_X(x)\}.
\end{equation}

The state-of-the-art quantile estimation using asymmetric signum functions is studied in~\cite{qt11} and~\cite{qt17}. The quantile approximation calibrates the quantiles in a sequential manner according to every incoming sample. The quantile calibration process from sample $x_{t-1}$ to $x_t$ can be described as
\begin{equation}
\footnotesize
\label{eq: qtupdate}
Q_t(\alpha) =  Q_{t-1}(\alpha) - \lambda \sgn_{\alpha}(Q_{t-1}(\alpha)-x_t),
 \end{equation}
where $\sgn_{\alpha}(\cdot)$ is the asymmetric signum function defined by
\begin{equation}
\footnotesize
\label{eq:sn}
\sgn_{\alpha}(z) = \begin{cases}
- \alpha, \quad \iif\ z < 0\\
              1 - \alpha, \quad \iif\ z \geq 0
            \end{cases}.
\end{equation}

\vspace{-3mm}
\subsection{Learning Numeric Attributes with Quantile Approximation}
\label{ssec: l_n_quantile}

To handle numeric attributes, we develop a new algorithm in the Hoeffding tree induction process by applying the quantile estimation with asymmetric signum functions, which is described in Algorithm~\ref{alg: qtht}. The proposed algorithm encompasses two key features: 1) a separate set of quantiles is maintained per attribute per class (line 6 to 12); and 2) the strategy to get left/right partitions based on the attribute distributions (line 14 to 22) has been customized to support the quantile method. Note that the number of quantiles to use is determined by the characteristics of the datasets. This is studied in Section~\ref{subsec: qtnum}.

A straightforward method~\cite{vfml} to deduce the partitions is to view each sample as a split point and compute distribution individually: for an attribute $i$ and a specific sample's attribute as the split point $pt_i$, an arbitrary sample is sorted to the left partition if its attribute value $a_i \leq pt_i$, or otherwise, it is filtered to the right partition. In our algorithm, we learn the samples with quantiles and represent sample distribution in CDF: each quantile value $Q({\alpha}_k)$ indicates that the percentage is ${\alpha}_k$ for the samples with the attribute values smaller than $Q({\alpha}_k)$. In this way, sample storage is not required. 

\begin{figure}[t]
	\begin{center}
		\includegraphics[width=0.75\linewidth]{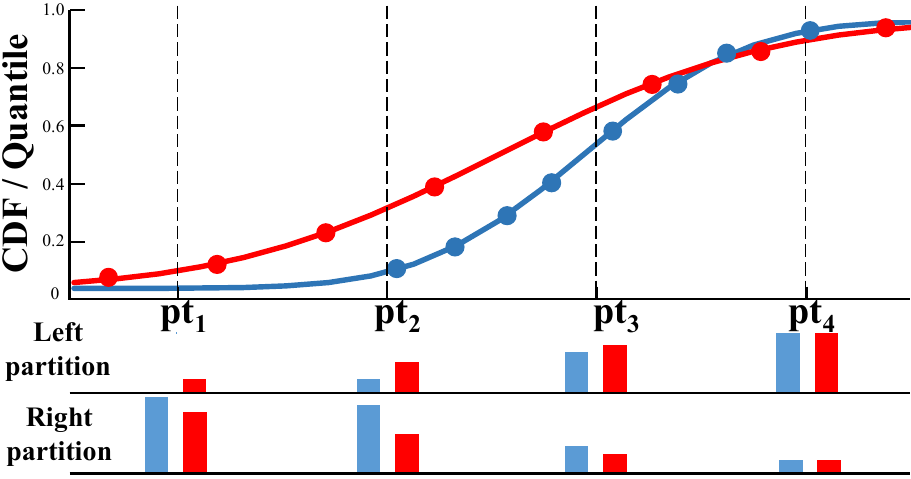}
		\vspace{-2mm}
		\caption{Partition strategy in the proposed algorithm, illustrated with one attribute, two labels and eight quantiles.}
		\label{fig: dist}
	\end{center}
\end{figure}

\begin{figure}[t]
	\begin{center}
		\includegraphics[width=0.6\linewidth]{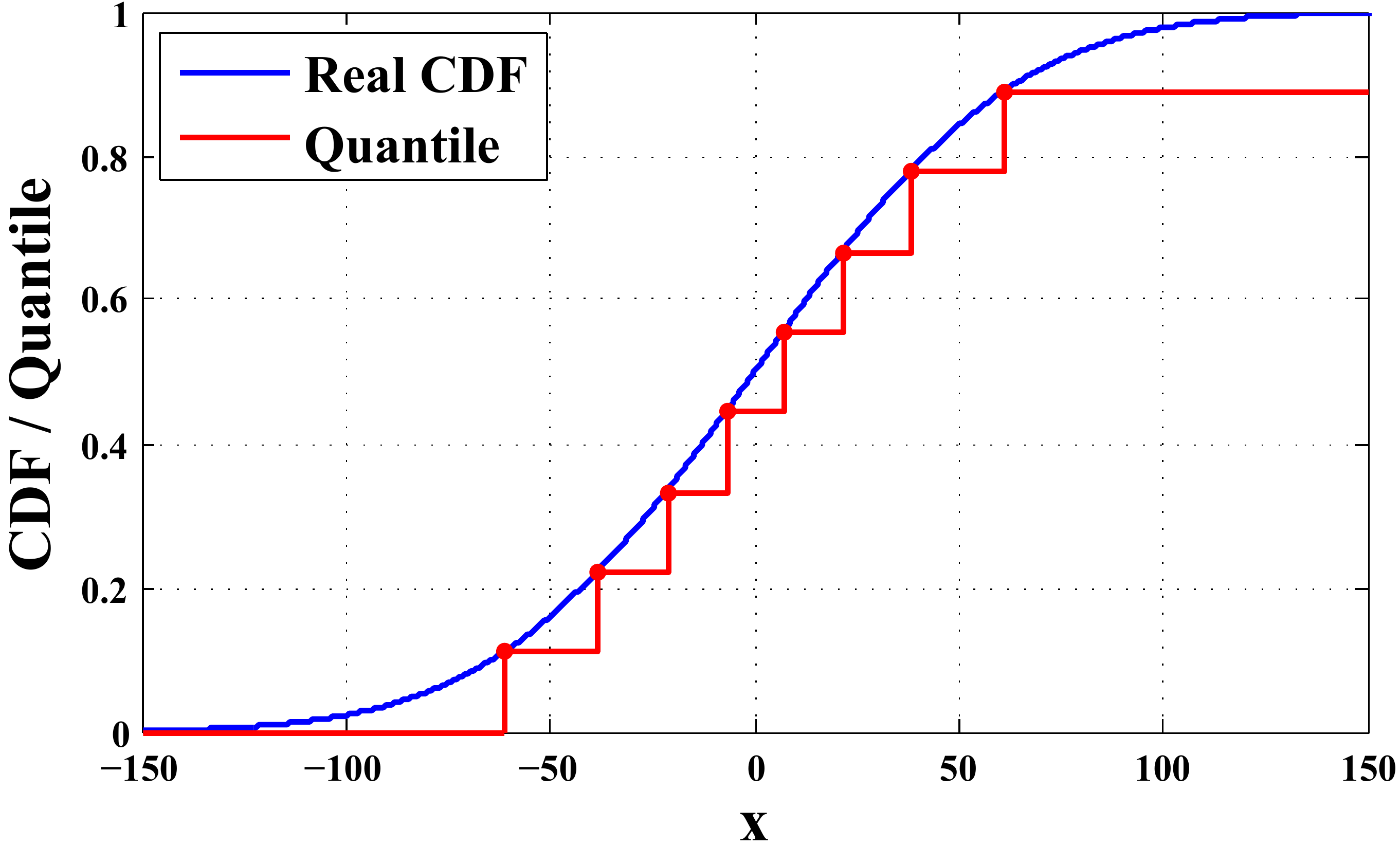}
		\vspace{-3mm}
		\caption{Using eight quantiles to estimate the CDF of normal distribution with a round-down scheme.}
		\label{fig: qt}
	\end{center}
\end{figure}

Fig.~\ref{fig: dist} illustrates how the overall partitioning strategy works. We generate a set of split points evenly distributed in the full range of attribute values. These split points are compared to the quantiles individually to find out the interval of two quantiles [$Q({\alpha}_k)$, $Q({\alpha}_{k+1})$] containing the split point. Afterwards, the sample number in each partition can be determined. The portion of samples with attribute values smaller than or equal to $Q({\alpha}_k)$ goes to the left partition, whereas the others go to the right partition. By this method, the sample distribution in the left partition is rounded down to the nearest quantile, with an example shown in Fig.~\ref{fig: qt}.

The proposed algorithm overcomes the trade-off between memory and computation, and presents a more rational and balanced solution compared with state-of-the-art methods~\cite{vfml, vfdtc, ga08, streamdm}. The advantages of this proposed method are three-fold. Firstly, the sample characteristics are fully generalized and encapsulated in a set of quantiles, dispensing with the need to store any samples in the training iterations. The memory requirement is reduced to $\#attribute \times \#label \times \#quantile$ per leaf node. This outperforms existing methods~\cite{vfml, vfdtc} which require large attribute or sample storage. Secondly, the computation demand is notably reduced compared with the memory-efficient yet computation-intensive method, Gaussian method~\cite{ga08, streamdm}: only comparison and subtraction are involved in quantile approximation, whereas Gaussian approximation entails expensive computation as shown in Algorithm~\ref{alg: gausapp}. The complexity of partition deduction is also effectively simplified with the proposed method. Thirdly, the problem of data dependency can be resolved with hardware optimization through deliberate parallelism and pipelining, as introduced in Section~\ref{subsec: lna}.

\section{Architecture Design}
\label{sec: impl}
\subsection{System Overview}
\begin{figure}[t]
	\begin{center}
		\includegraphics[width=0.65\linewidth]{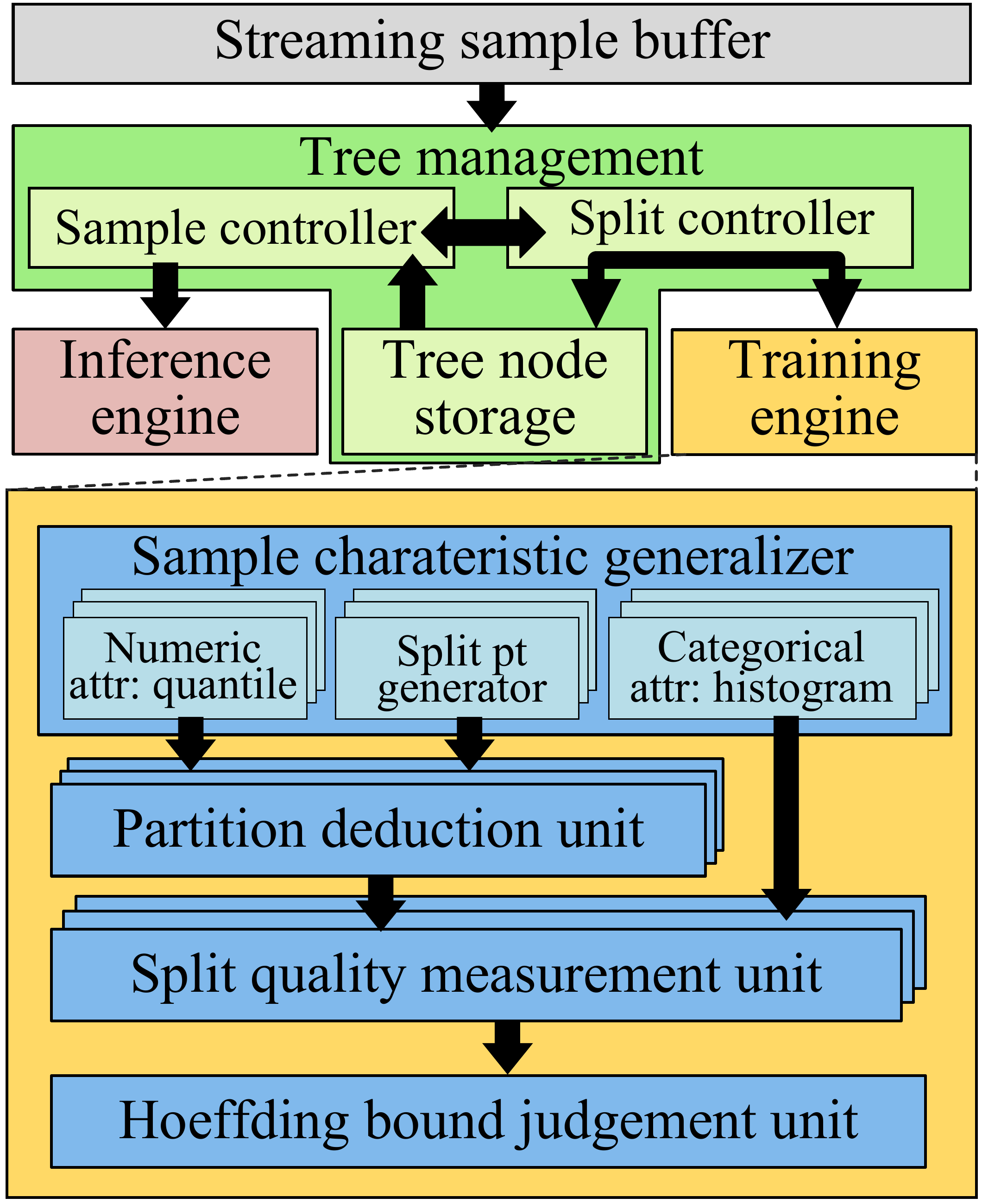}
		\vspace{-3mm}
		\caption{System overview of the Hoeffding tree implementation with the proposed algorithm.}
		\label{fig: arch}
	\end{center}
\end{figure}

The system overview of the Hoeffding tree implementation is depicted in Fig.~\ref{fig: arch}. Starting from the sample buffer, the tree management engine first reads and decodes the sample information. At the same time, it fetches relevant tree nodes from the tree node storage and filters the samples to the leaf nodes in a pipelined way. Thereafter, both the inference engine and training engine start processing the samples.

In the learning process, samples are decomposed into separate attributes and the characteristics of each attribute are learned and stored independently. When a split trial is invoked at a leaf node, for each numeric attribute, a partition deduction unit uses the quantiles and split points to deduce left and right partitions. As for categorical attributes, the sample counts of all attribute-class combinations form a histogram, which is similar to the quantiles for numeric attributes. 

The partition information of every attribute is then processed by a split quality measurement unit to compute the split gain for each split point. Then, the best and second-best split gains are identified, and the split decision is given by the Hoeffding bound judgment unit. If a split is taken, the split information is sent back to the split controller to update the tree structure.

\vspace{-3mm}
\subsection{Tree Management Units}
\label{ssec: tree}
The tree management units maintain two operations: 1) filtering samples to different leaf nodes, which requires tree traversing; and 2) splitting leaf nodes by overwriting the tree node memory after receiving split requests.

\begin{figure}[t]
	\begin{center}
		\includegraphics[width=0.9\linewidth]{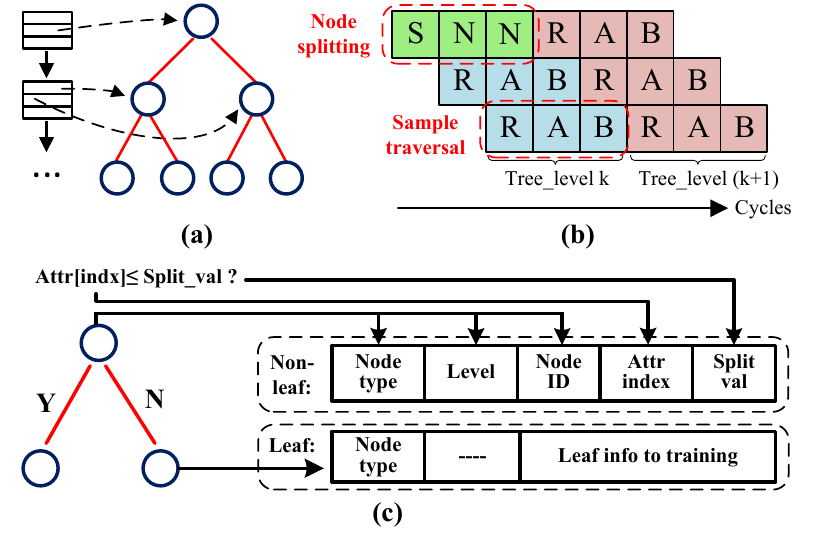}\\
		\vspace{-3mm}
		\caption{(a) Decision tree architecture; (b) tree management pipeline structure; and (c) bit decomposition of tree node memory.}
		\vspace{-2mm}
		\label{fig: dt}
	\end{center}
\end{figure}

The tree traversing process for each sample starts from the root node down to a specific leaf node, thus involving several rounds of memory reading. Considering the case of streaming data input, the tree memory may receive multiple read requests from different samples concurrently. Multi-port memory can be used to support this feature. However, the required port number is linearly related to the tree depths. FPGA BRAMs naturally support up to two ports, and increasing the port size turns out to be an inefficient solution. We observe that the samples are processed at different tree levels sequentially and the samples from different time steps require memory reading from different tree levels. Hence, we separate the node storage according to tree levels, as depicted in Fig.~\ref{fig: dt} (a), and dual-port memory is enough to support both node splitting and tree traversing for streaming samples. The idea of using a separate memory structure has been adopted in DT-CAIF~\cite{saq15}, whereas we develop a fine-grained pipeline structure for each tree level. All the tree levels together form a deep pipeline.

The fine-grained pipeline needs to support both tree traversing and node splitting. A three-stage pipeline is formed, as shown in Fig.~\ref{fig: dt} (b). The tree traversing routine consists of node reading (R), attribute selection (A) and branch decision (B) stages. As for node splitting, split information (mainly the split node level, node ID, split coefficient and attribute index) from the training engine is passed across different tree levels. When a leaf node is reached, the corresponding memory element is overwritten by the split information to replace the leaf node with an internal node. Moreover, two new leaf nodes are generated in the next level and the split pipeline also writes in the new leaf nodes the training elements they are associated to. This is related to the dynamic leaf node-element allocation scheme discussed in Section~\ref{subsec: lna}. All the operations relevant to the split are completed in the split (S) stage, after which two nop (N) states are followed. The bit information stored in the memory for branch decision is shown in Fig.~\ref{fig: dt} (c). 

\vspace{-2mm}
\subsection{Learning Numeric Attributes}
\label{subsec: lna}
In our proposed Algorithm~\ref{alg: qtht}, recall that we maintain a set of quantile values per numeric attribute per class for a single leaf node. Optimization techniques are investigated for accelerating quantile learning from the hardware perspective, which can be summarized as: 1) attribute-level (coarse-grained) and quantile-level (fine-grained) parallelism; 2) dynamic and memory-based resource sharing; and 3) pipelining with data forwarding for data dependency removal.

\begin{figure}[t]
	\begin{center}
		\includegraphics[width=0.9\linewidth]{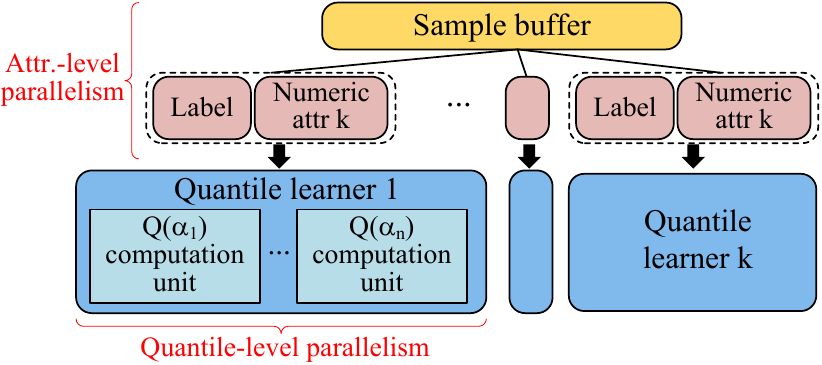}
		\vspace{-3mm}
		\caption{Exploiting attribute-level and quantile-level parallelism.}
		\label{fig: qtpar}
	\end{center}
\end{figure}

\textbf{Attribute-level (Coarse-grained) and Quantile-level (Fine-grained) Parallelism.} As shown in line 6 to line 12 of Algorithm~\ref{alg: qtht}, different attributes are independent and, within each attribute, the quantiles $Q(\cdot)$ per class are also independent of each other. This allows us to speed up the quantile computation process with both attribute-level and quantile-level parallelism, as shown in Fig.~\ref{fig: qtpar}. Note that we do not take class-level parallelism even though it is possible. This is because each sample contains a unique class label but has multiple attributes. The learning process only needs to update the set of quantiles matching the sample label. Based on this fact, parallelizing at class level does not offer any benefit. Instead, we seek opportunities for class-level optimization through resource sharing and pipelining.

\textbf{Dynamic and Memory-based Resource Sharing.} For each leaf node, it is required to maintain a number of quantiles per attribute per class. If hardware copies are simply replicated for each leaf node, both the memory and arithmetic resource utilization becomes too expensive for hardware to implement. In light of this problem, we develop a dynamic leaf node-element allocation scheme as the tree grows dynamically and a memory-based resource sharing mechanism for quantile update routine. 

\begin{figure}[t]
	\begin{center}
		\vspace{-2mm}
		\includegraphics[width=\linewidth]{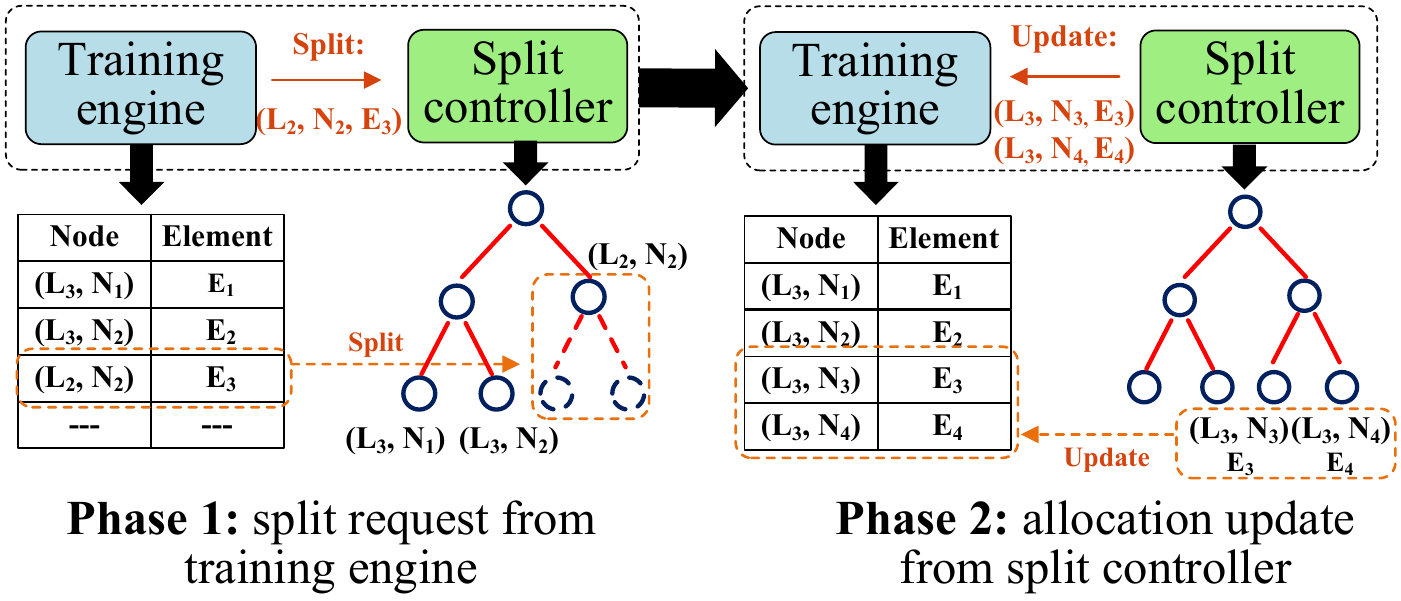}
		\vspace{-7mm}
		\caption{Dynamic leaf node-element allocation scheme.}
		\vspace{-2mm}
		\label{fig: dyn}
	\end{center}
\end{figure}

To differentiate between a leaf node of the tree and the physical resource allocated for a leaf node in the training process, we call the former a \emph{leaf node}, while we denote the latter as an \emph{element}. A leaf node is only temporarily being a leaf node, and it may be split as samples assemble. Therefore, it is not necessary to statically allocate physical resources to each leaf node. We devise a \emph{dynamic leaf node-element allocation scheme}, as shown in Fig.~\ref{fig: dyn}. The training engine maintains a node-element table to keep track of the leaf node-element pairs. During the split process, the split controller generates new leaf node-element pairs and sends them back to the training engine. The traning engine then updates the leaf node-element relationship in the table. In this way, the leaf node-element allocation change dynamically and resource reuse in hardware is facilitated.

\begin{figure}[t]
	\begin{center}
		\includegraphics[width=0.7\linewidth]{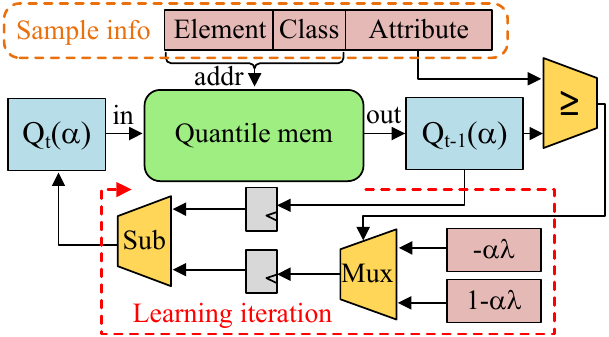}
		\vspace{-3mm}
		\caption{A single quantile computation unit with the memory-based resource sharing scheme.}
		\label{fig: mbqta}
	\end{center}
\end{figure}

A \emph{memory-based resource sharing scheme} is designed to collaboratively work with the dynamic leaf node-element allocation scheme for further resource sharing. This scheme leverages two facts: 1) each sample is only sorted to one leaf node, so only one element will be activated for quantile update per sample; and 2) for each attribute, only the set of quantiles corresponding to the sample label will be activated per sample. Since the quantile learning process is the same for all classes and elements, except that the quantile values are different, we devise the following memory-based resource sharing scheme: for each attribute, all the classes of all elements share one set of quantile computation logics and all the corresponding quantile values are stored in one memory. When a sample is used for training, the set of quantiles corresponding to the specific element and class is fetched, and later, the updated values are stored back to the same memory location. Element and class values together form the memory addresses. Putting it all together, a single quantile computation unit with memory-based resource sharing is depicted in Fig.~\ref{fig: mbqta}. To support this mechanism, each leaf node in the tree memory preserves a field denoted as \emph{leaf information to training} shown in Fig.~\ref{fig: dt} (c). Provided a new split, the two new leaf nodes along with their assigned element IDs are sent from the training engine to the tree node memory for update. For each sample after tree traversing, the element ID associated with its reached leaf node and the raw sample data are sent to the training engine.

\begin{figure}[t]
	\begin{center}
		\vspace{-3mm}
		\includegraphics[width=0.9\linewidth]{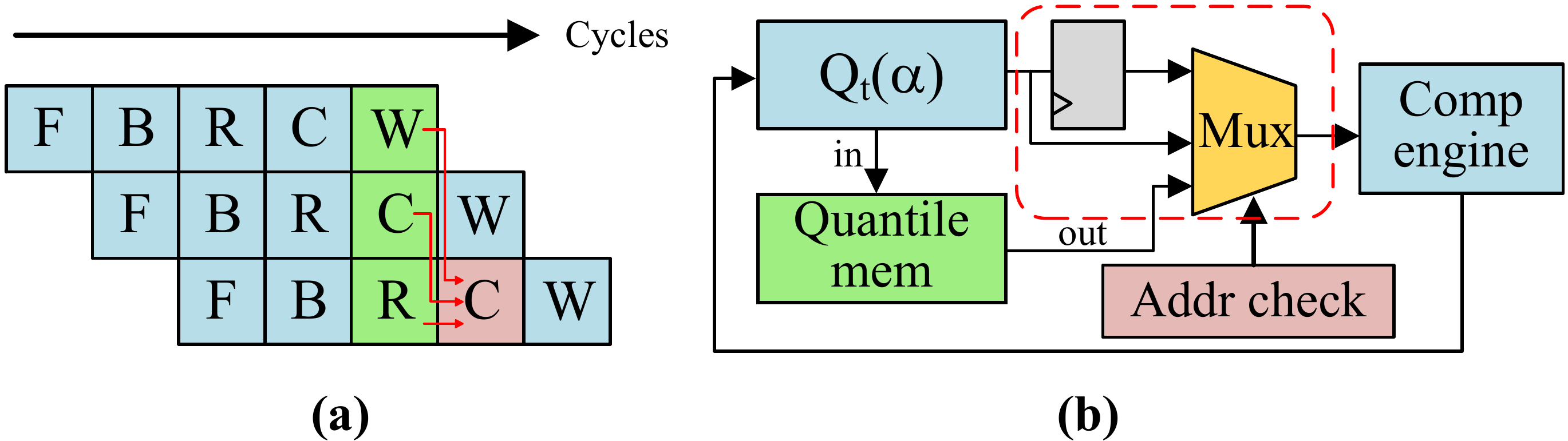}
		\vspace{-3mm}
		\caption{(a) Pipelining stages with data forwarding; (b) hardware realization of data forwarding.}
		\label{fig: qtpipe}
	\end{center}
\end{figure}

\textbf{Pipelining with Data Forwarding.} There exists data dependency for quantile computation: two successive samples sorted to the same leaf node should update the same element in a sequential way. For the method with Gaussian approximation, the long latency of the update process, as described in Algorithm~\ref{alg: gausapp}, makes it difficult to overcome this dependency. For the proposed quantile computation architecture in Fig.~\ref{fig: mbqta}, the computation is reduced to a comparison and a subtraction per quantile unit, which allows us to fully exploit the pipeline architecture with data forwarding to resolve data dependency. 

We propose a 5-stage pipeline architecture for the quantile update routine, as shown in Fig.~\ref{fig: qtpipe} (a). The first stage (F) fetches a sample from the sample buffer. The second stage (B) decides on the execution branch to take, including element initialization in the dynamic leaf node-element allocation scheme, quantile computation and quantile output for the split trial. In the next stage (R), the quantile unit selected by the element and class is read out. Afterwards, the quantile is updated in the computation stage (C) following Equation~(\ref{eq: qtupdate}), and is written back to the same memory location in the writing stage (W).

In stage C, we address the data dependency problem by the adoption of a dedicated data forwarding method, as shown in Fig.~\ref{fig: qtpipe} (b), which aims at providing the flexibility that, when the quantiles are updated while not yet written in the memory, they are directly passed to the quantile computation engine if the addresses between these two computation periods match. We keep track of the results and quantile memory addresses of the prior two computation periods, which are managed by stage C and stage W, respectively. Stage C has a higher forwarding priority over stage W when both memory addresses match the one currently processing, because stage C provides the most up-to-date results. This data forwarding allows us to bypass memory operations when dependency occurs and eventually leads to a throughput of one sample per cycle.

\vspace{-3mm}
\subsection{Learning Categorical Attributes}
\label{ssec: learn_categ}

The process of learning categorical attributes is similar to learning numeric attributes. 
However, the value and size of each categorical attribute is determined by dataset characteristics, which can be known in design time. Therefore, counting the number of occurrence for each attribute-class combination gives a histogram of the distribution without any loss of information. In a split trial for categorical attributes, each attribute value is used as a split point individually: the samples with the attribute value equal to the split point is filtered to the left, or otherwise, it is sorted to the right.

\begin{figure}[t]
	\begin{center}
		\includegraphics[width=0.9\linewidth]{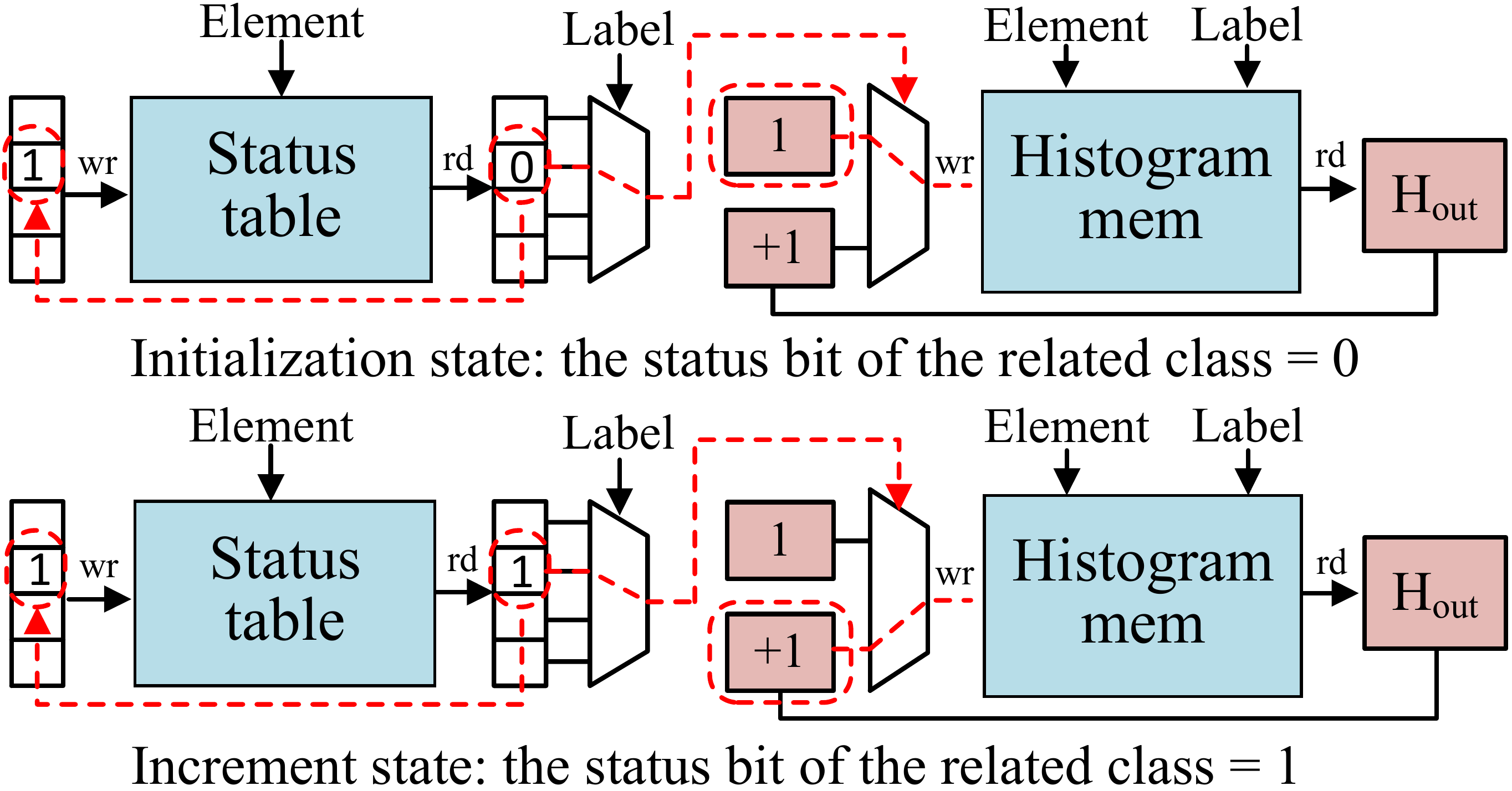}
		\vspace{-2mm}
		\caption{Histogram update with status table.}
		\label{fig: hist}
	\end{center}
\end{figure}

The optimization methods, except the dynamic leaf node-element allocation scheme, can be migrated to categorical attributes seamlessly. However, to support the dynamic leaf node-element allocation scheme, the histograms of all attribute-class combinations for an element need to be initialized simultaneously. This brings difficulties as we apply memory-based resource sharing in which the same dual-port histogram memory is shared amongst different class labels, and multiple write requests to the same memory is inefficient for FPGA design. To overcome this problem, we additionally implement a status table for histograms. Every memory unit in the status table represents an individual histogram, and each bit indicates the status of a column of this histogram. 
To initialize a histogram when a new leaf node-element pair is assigned, only the corresponding memory unit in the status table, instead of all units in the histogram, needs to be reset. The training routine first checks the status table for each incoming sample, and follows either of the two situations (i.e., initialization or increment) as depicted in Fig.~\ref{fig: hist}. The relevant status bit is set to high when the first sample comes after initialization.

\vspace{-3mm}
\subsection{Simplification of Split Measurement with Hoeffding Bound}
The study in~\cite{tan07} has shown that the choice of split measurement method does not exert a significant impact on the accuracy of decision tree induction. We adopt gini impurity~\cite{cart} as it is commonly used and has low computational demand.

Gini impurity is a measure of the chance for an example to be incorrectly classified if it is randomly labeled according to the distribution of the labels. Let $p_j$ be the probability of examples being labeled as class $j\ (j \in{1,2,...,|C|})$ in the dataset $S$. Gini impurity can be represented as
\begin{equation}
\footnotesize
\label{eq: gini}
gini(S) = 1-\sum_{j=1}^{|C|}p_j^2.
\end{equation}
The split quality for a given partition is based on the reduction in gini impurity after a split is taken. If $S$ is split into the left subset $S_L$ and right subset $S_R$, the reduction in gini impurity can be described by
\begin{equation}
\footnotesize
\label{eq: ginigain}
G = \Delta gini = gini(S) - \frac{|S_L|}{|S|}gini(S_L) - \frac{|S_R|}{|S|}gini(S_R).
\end{equation}

We combine the split measurement with the Hoeffding bound judgment for joint optimization in hardware. Let $S_{L, j}$ and $S_{R, j}$ be the subset of $S_L$ and $S_R$ labeled in $j$, respectively. We reorganize the reduction in gini impurity $G$ as follows:
\begin{equation}
\footnotesize
\label{eq: reginigain}
G = \frac{1}{|S|}(\frac{1}{|S_L|}\sum_{j=1}^{|C|}|S_{L,j}|^2 + \frac{1}{|S_R|}\sum_{j=1}^{|C|}|S_{R,j}|^2) + gini(S)-1.
\end{equation}
Putting the gini impurity reduction and Hoeffding bound together, the calculation can be reorganized as
\begin{equation}
\scriptsize
\begin{aligned}
\label{eq: ginihb}
G_{B_1}-G_{B_2} = & \frac{1}{|S|} \big[ \underbrace{(\frac{1}{|S_{B_1,L}|}\sum_{j=1}^{|C|}|S_{B_1,L,j}|^2 +  \frac{1}{|S_{B_1,R}|}\sum_{j=1}^{|C|}|S_{B_1,R,j}|^2)}_{\textnormal{split quality}} - \\& (\frac{1}{|S_{B_2,L}|}\sum_{j=1}^{|C|}|S_{B_2,L,j}|^2 + \frac{1}{|S_{B_2,R}|}\sum_{j=1}^{|C|}|S_{B_2,R,j}|^2) \big].
\end{aligned}
\end{equation}
To search for the best and second-best attributes, we only need to compute the \emph{split quality} denoted in Equation~(\ref{eq: ginihb}) for each split point, instead of the full term of gini impurity reduction in Equation~(\ref{eq: reginigain}). After that, the whole term of Equation~(\ref{eq: ginihb}) is computed for Hoeffding bound judgment. This noticeably simplifies the calculation for each split point.

To further optimize the computation, we eliminate the division $\frac{1}{|S|}$ in Equation~(\ref{eq: ginihb}) by pre-storing and looking up the values in memory. The square-sum calculation in the split quality term is realized with a pipelined multiplier-adder tree.

\section{Performance Model}
\label{sec: odt_pm}
In this section, we analyze and model the performance metrics of the online decision tree. These principal metrics are inference latency, throughput and execution time. The inference latency describes the execution cycles for a sample to pass through the inference engine and get prediction results. The throughput reflects the total data volume that is processed per unit time. The execution time represents the overall system operation time for an application with a dataset.

\vspace{-5mm}
\subsection{Inference Latency}
\label{subsec: inf_lat}
The latency measurement starts from a single input sample already transferred to the FPGA to the point where the inference results are ready to be sent out to off-chip memory from the FPGA. The overall inference latency $L_{overall}$ encompasses three major components: buffer delay $L_{buff}$, tree traverse delay $L_{tree}$, and prediction delay $L_{pred}$. Eq.~\ref{eq: lat_overall} describes this relationship:
\begin{equation}
\footnotesize
\label{eq: lat_overall}
L_{overall} = L_{buff} + L_{tree} + L_{pred}.
\end{equation}

The buffer delay $L_{buff}$ accounts for two buffers, the input sample buffer and the internal data buffer between the tree traverse unit and the prediction unit. Each buffer takes four cycles for the data to be stored and outputted according to the profiling results of the Xilinx FIFO IP block~\cite{xfifoip}. Therefore, $L_{buff}$ is eight in our case. The tree traverse delay $L_{tree}$ corresponds to the latency incurred by sorting a sample from the root node down to a leaf node in the tree body. As described in Section~\ref{ssec: tree}, the tree traverse unit is customized with both intra-level and inter-level pipeline. Putting it all together, $L_{tree}$ can be formulated as
\begin{equation}
\footnotesize
\label{eq: lat_tree}
L_{tree} = P_{level} \times D_{tree},
\end{equation}
where $P_{level}$ denotes the number of pipeline stages per tree level, which are three according to Section~\ref{ssec: tree}, and $D_{tree}$ is the user-defined maximum tree depth. The prediction delay $L_{pred}$ describes the delay to provide inference results given the sample and leaf node information after the tree traverse. As majority vote is adopted as the inference strategy, the data count per label and maximum data count per leaf node are maintained and updated for each leaf node. Accordingly, a three-stage pipeline is designed for this purpose with read, predict and update stages deployed in order. The read stage fetches the corresponding label and maximum data counts of the leaf node from the memory, the predict stage provides the inference result based on the majority vote strategy, and finally, the update stage conducts current label count and maximum label count increment. According to this pipeline, the prediction unit offers inference results at the second pipeline stage, and it takes two cycles for result outputting per sample, and therefore, $L_{pred}$ is two.

\vspace{-4mm}
\subsection{Throughput}
Our design is fully pipelined and the design is able to process one sample per cycle. Under this situation, the full-load throughput of the FPGA $TP_{FPGA}$ is formulated as
\begin{equation}
\footnotesize
\begin{gathered}
\label{eq: tp}
TP_{FPGA} = \left[ \log_2(L) + B_N \times N + B_C \times C \right] \times f, \\
\text{with} \quad B_N = 32, \ \ 
B_C = \max\{\left \lceil{\log_2(V_i)}\right \rceil : 1 \leq i \leq C\},
\end{gathered}
\end{equation}
where the first term of $TP_{FPGA}$ is the sample size that is mainly determined by the number of numeric attributes $N$, the number of categorical attributes $C$ and the number of labels $L$ in the target application, and $f$ is the operating frequency given after the application going through logic synthesis, placement and routing. $B_N$ and $B_C$ are the bitwidths of numeric and categorical attributes, respectively. The numeric attributes are 32-bit fixed-point data in our design, so $B_N$ is 32. $B_C$ is the bitwidth required to cover the range of all categorical attribute values, which is computed as shown in Eq.~\ref{eq: tp}, with $V_i$ representing the number of attribute values for the $i$th categorical attribute.

Besides the design throughput, the overall throughput should also be responsible for the DDR bandwidth. The measured DDR bandwidth is reported as 9.5 GB/s for reading (denoted as $DBW_{RD}$) and 8.9 GB/s for writing (denoted as $DBW_{WR}$)~\cite{choi16}. In our case, the bottleneck of DDR accessing is caused by reading samples. Finally, the overall throughput $TP_{overall}$ of the complete system is given by
\begin{equation}
\footnotesize
\label{eq: tpoverall}
TP_{overall} = \min(TP_{FPGA},\ DBW_{RD}).
\end{equation}

\vspace{-5mm}
\subsection{Execution Time}
The overall execution time $T_{exe}$ for an application can be generalized as
\begin{equation}
\footnotesize
\label{eq: texe}
T_{exe} = \frac{C_{exe}}{f},
\end{equation}
where $C_{exe}$ is the execution cycles of the target application given a specific dataset, and $f$ is the operating frequency. The execution cycles can be described by
\begin{equation}
\footnotesize
\label{eq: cexe}
C_{exe} = C_{per-sample} \times S + C_{cold-start},
\end{equation}
where $C_{per-sample}$ is the amortized execution cycles for each sample, $S$ denotes the sample size of the used dataset, and $C_{cold-start}$ represents the execution cycles used to bring the design pipeline into its normal operation at the system startup stage. During profiling in the experiments, $C_{per-sample}$ is observed to be 1.047, which is close to one. This corresponds to the fact that our design is fully pipelined with an initiation interval (II) equal to one, capable of processing a new input every clock cycle. $C_{cold-start}$ corresponds to the following phases at the system startup stage: 1) data reading, buffering and writing; 2) pipeline filling and draining; 3) sample accumulation at leaf nodes, split decision feedback and update.

\vspace{-3mm}
\section{Resource Model}
\label{sec: odt_rm}
We introduce a resource model to investigate the relationship between application characteristics and the corresponding resource overheads. This model can help designers understand the resource decomposition of different components in the design, and it also offers fast and early-stage resource evaluation simply based on some application parameters. As DSPs and BRAMs are scarce resources for FPGA and the hardware implementation of online decision trees is both memory and computation intensive, we focus on modeling the DSP and BRAM resource utilization. 

\vspace{-3mm}
\subsection{DSP Utilization}
We assume that all multiplications use the DSPs for high performance. In this design, we use integer multipliers with 16-bit, 24-bit and 32-bit configurations for the fixed-point multiplications in the design, each consuming one, two, and four DSP slices, respectively. The DSPs are utilized by 1) the numeric attribute learning; 2) the categorical attribute learning; and 3) the split decision, as formulated in Eq.~\ref{eq: dsp_overall}:
\begin{equation}
\footnotesize
\label{eq: dsp_overall}
D_{overall} = D_{numeric} + D_{categorical} + D_{split}.
\end{equation}

\textbf{DSP in Numeric Attribute Learning.} For numeric attribute learning, a number of multipliers are used in the computation of Gini impurity after several split attempts are provided, which is denoted as the \textit{split quality} term in Eq.~\ref{eq: ginihb}. Regarding each attribute, the multiplications are involved in 1) squaring the sample counts per label in both the left and the right partitions, and 2) normalizing the left and the right square sums with the partition sum, respectively. In light of this, the DSP utilization for numeric attribute learning $D_{numeric}$ is given by
\begin{equation}
\footnotesize
\label{eq: dsp_n}
D_{numeric} = (3 \times L + 12) \times N,
\end{equation}
where $L$ is the number of labels, and $N$ is the number of numeric attributes. From Eq.~\ref{eq: dsp_n}, we can observe that the DSP utilization for numeric attribute learning is determined by the number of labels and the number of numeric attributes, which are intrinsically decided by the application characteristics.

\textbf{DSP in Categorical Attribute Learning.} For categorical attribute learning, the multiplications are similarly utilized in the computation of the split quality term. The main difference between the numeric attribute learning and the categorical attribute learning is the representation of sample distribution per attribute-class combination: for numeric attribute learning, the sample distribution is learned in the form of quantiles, and 32-bit fixed-point representation is used to maintain the precision, whereas for categorical attribute learning, the sample distribution is preserved in a histogram which records the occurrence of samples, and 16-bit integer is enough to cover a large range. This contributes to the differences in the DSP utilization for the computation of the same split quality term for numeric and categorical attribute learning. Accordingly, the DSP utilization for categorical attribute learning $D_{categorical}$ is generalized as
\begin{equation}
\footnotesize
\label{eq: dsp_c}
D_{categorical} = (2 \times L + 4) \times C,
\end{equation}
where $C$ represents the number of categorical attributes. Similar to numeric attribute learning, the DSP utilization for categorical attribute learning is also determined by the number of labels and the number of categorical attributes in the dataset.

\textbf{DSP in Split Decision.} For split decision, a 32-bit multiplier is used to multiply the normalization term $\frac{1}{|S|}$ with the subtraction result of split quality terms, as shown in Eq.~\ref{eq: ginihb}. Therefore, four DSPs are used for split decision after the Gini computation, so $D_{split}$ equals to four.

\vspace{-3mm}
\subsection{BRAM Utilization}
\label{subsec: bram}
The online decision tree requires intensive usage of BRAM resources in both the inference and the training processes. We decompose the complete design into inference, numeric attribute learning and categorical attribute learning, and separately study their BRAM utilization. The BRAM utilization model is constructed according to the state-of-the-art Xilinx Ultrascale+ FPGA BRAM features~\cite{xbram}.

\textbf{BRAM in Inference.} In the inference stage, the BRAM resources are modeled by
\begin{equation}
\footnotesize
\label{eq: bram_inf}
B_{inference} = B_{buff\_i} + B_{tree} + B_{pred},
\end{equation}
where $B_{buff\_i}$, $B_{tree}$ and $B_{pred}$ denote the BRAMs utilized by data buffering, tree traverse and prediction, respectively. 

$B_{buff\_i}$ consists of input and internal buffers with RAMs and FIFOs. These are a small portion in $B_{inference}$, and can be regarded as a constant in the design. The $B_{buff\_i}$ is observed to be 32 through profiling. 

$B_{tree}$ represents the memory used for tree node storage. To increase the efficiency of pipelining in the design, we allocate one individual memory for each tree level, and the overall utilization is to add up the BRAMs utilized for different components, as given by
\begin{equation}
\footnotesize
\label{eq: b_tree}
B_{tree} = \sum_{level = 1}^{D_{tree}} B_{level},
\end{equation}
where $D_{tree}$ denotes the maximum tree depth that is set by users, and $level$ is the currently evaluated tree level. The number of tree nodes for different levels are different, which can be generalized as
\begin{equation}
\footnotesize
\label{eq: lsize}
Node(level) = 2^{level - 1}.
\end{equation}
Due to this observation, the BRAM utilization also differs for different tree levels. We separately formulate the BRAM utilization according to specific tree levels:
\begin{equation}
\footnotesize
\label{eq: b_level}
	B_{level} = \begin{cases}
	0, \quad \iif\ level = 1 \\[10pt]
	\left \lceil{
	\frac{33 + D_{tree} + 
		\left \lceil{\log_2(D_{tree})}\right \rceil + \left \lceil{\log_2(C+N)}\right \rceil
	}{18}}
	\right \rceil
	, 
	\iif\ level \in [2, 11] \\[10pt]
	
	\left \lceil{
	\frac{33 + D_{tree} + 
		\left \lceil{\log_2(D_{tree})}\right \rceil + \left \lceil{\log_2(C+N)}\right \rceil
	}{36}}
	\right \rceil
	\times 2^{level-11} + 4
	,\\ \qquad \qquad \qquad \qquad \qquad \qquad \qquad \qquad \qquad \qquad \iif\ level \geq 12
\end{cases}
\end{equation}
where $C$ is the number of categorical attributes and $N$ is the number of numeric attributes. For the first level, there is only one root node, so only registers are used to simply buffer the root node and no BRAM memory is required. For the levels between 2 and 11, the number of nodes is no larger than 1024. Under this circumstance, the BRAM utilization is determined by the data width. In contrast, when the tree level further increases, the BRAM utilization is also influenced by the data size, namely, the number of nodes in the tree level.

$B_{pred}$ can be decomposed into sample label storage of leaf nodes, and majority class update and memorization for leaf nodes, as shown in Section~\ref{subsec: inf_lat}. $B_{pred}$ can be described as
\begin{equation}
\footnotesize
\begin{aligned}
\label{eq: b_pred}
B_{pred} = &
\left \lceil{
	2^{\left \lceil{\log_2(E)}\right \rceil + \left \lceil{\log_2(L)}\right \rceil - 10}
}\right \rceil + \left \lceil{\frac{E}{1024}}\right \rceil
\\& + \left \lceil{
		\frac{12+L+\left \lceil{\log_2(L)}\right \rceil}{18}}\right \rceil \times \left \lceil{\frac{E}{1024}
	}\right \rceil 
.
\end{aligned}
\end{equation}

\textbf{BRAM in Numeric Attribute Learning.} BRAMs are extensively used in numeric attribute learning for 1) internal data buffering, and 2) quantile learning, as shown in Eq.~\ref{eq: bram_n}:
\begin{equation}
\footnotesize
\label{eq: bram_n}
B_{numeric} = B_{buff\_n} + B_{quantile}.
\end{equation}

There are several internal RAMs and FIFOs in numeric attribute learning for the purpose of internal data storage, ranging from sample storage, element status preservation, attribute/label range capturing to coefficient buffering, etc. The overall BRAM utilization for data buffering can be summarized as 
\begin{equation}
\footnotesize
\label{eq: b_n_buff}
\begin{split}
B_{buff\_n} = N \times \left( 8 + 2L + 
	\left \lceil{
		\frac{\left \lceil{\log_2(E)}\right \rceil + 
		\left \lceil{\log_2(L)}\right \rceil + 32
		}{18}
	}\right \rceil \right)
	\\ + N \times \left \lceil{\frac{E}{1024}}\right \rceil \times \left( 5  + 4L + \left \lceil{\frac{1+6L}{18}}\right \rceil + \left \lceil{\frac{2L}{3}}\right \rceil \right),
\end{split}
\end{equation}
where $N$, $L$ and $E$ denote the number of numeric attributes, the number of labels in the target application and the number of elements in the hardware design, respectively.

Regarding the quantile estimation, a set of quantiles are maintained per attribute-class combination for each leaf node, as described in Section~\ref{ssec: l_n_quantile}. As a result, the quantile storage gives rise to a major proportion of the overall BRAM utilization in the design. Eq.~\ref{eq: b_n_quantile} models the BRAM utilization for the quantile learning:
\begin{equation}
\footnotesize
\label{eq: b_n_quantile}
B_{quantile} =  N \times \left( 
\left \lceil{
	2^{\left \lceil{\log_2(E)}\right \rceil + \left \lceil{\log_2(L)}\right \rceil - 10}
}\right \rceil \times 2Q + 
\left \lceil{\frac{8Q}{9}}\right \rceil \times L
\right),
\end{equation}
where $Q$ represents the adopted number of quantiles per quantile set in the design. The first term of Eq.~\ref{eq: b_n_quantile} describes the resources used for quantile initialization, storage and update. The second term represents the intermediate buffers to transfer the results between quantile learning and partition deduction.

\textbf{BRAM in Categorical Attribute Learning.} Different from the quantile learning for numeric attributes, the categorical attribute learning process instead adopts a histogram representation, as described in Section~\ref{ssec: learn_categ}. The total number of BRAMs used for categorical attribute learning is described by
\begin{equation}
\footnotesize
\label{eq: bram_c}
B_{categorical} = B_{buff\_c} + B_{histo}.
\end{equation}

Multiple buffers are allocated to store input and internal data for histogram update and partition deduction. The BRAM utilization for data buffering is given by
\begin{equation}
\footnotesize
\label{eq: b_c_buff}
B_{buff\_c} = 
\left( \left \lceil{\frac{2L}{3}}\right \rceil + 2 \right) \times \left \lceil{\frac{E}{1024}}\right \rceil + 10 \times C,
\end{equation}
where $C$, $E$ and $L$ denote the number of categorical attributes, the number of elements and the number of labels of the design, respectively.

\begin{figure}[t]
	\begin{center}
		\includegraphics[width=0.8\linewidth]{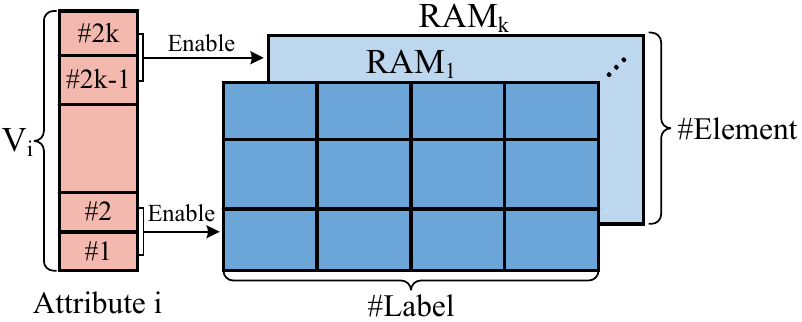}
		\vspace{-3mm}
		\caption{Buffer organization for histogram storage.}
		\label{fig: categ_ram}
	\end{center}
	\vspace{-1mm}
\end{figure}

To learn from categorical attributes, a histogram is maintained for each combination of attribute value and label in each leaf node. To enhance memory bandwidth for parallel data processing, we partition the memory in the dimension of attribute values, and allocate dual-port memory for every two attribute values, as shown in Fig.~\ref{fig: categ_ram}. The attribute values are used as enable signals to select the memory unit to access. Besides the memory utilization for histogram, there are also some buffers allocated to record the sum and status of histogram elements. Putting it all together, the BRAM utilization for histogram representation is 
\begin{equation}
\footnotesize
\begin{split}
\label{eq: b_c_histogram}
B_{histo} = \sum_{i=1}^{C} \sum_{j=1}^{\left \lceil{\frac{V_i}{2}}\right \rceil}
\left(
\left \lceil{
	2^{\left \lceil{\log_2(E)}\right \rceil - 9}
}\right \rceil + 
 3 \times \left \lceil{
	2^{\left \lceil{\log_2(E)}\right \rceil + \left \lceil{\log_2(L)}\right \rceil - 11}
}\right \rceil \right.
\\ \left. + \left \lceil{
	2^{\left \lceil{\log_2(E)}\right \rceil + \left \lceil{\log_2(L)}\right \rceil - 13}
}\right \rceil + 
\left \lceil{\frac{E}{8192}}\right \rceil
\right),
\end{split}
\end{equation}
where $V_i$ denotes the total number of attribute values for the $i$th categorical attribute in the application.

\section{FPGA Run-Time Power Monitoring with Online Learning}
In this section, we study how Hard-ODT, the online decision tree learning system proposed in prior sections of this paper, can be further utilized for FPGA run-time power monitoring. We note that state-of-the-art research works~\cite{lin17, lin18} have focused on offline power modeling strategies, which models the FPGA power consumption by collecting samples for training beforehand. The work~\cite{lin17} proposed a computer-aided design (CAD) flow to train decision tree models as power indicators, and devised a light-weight architecture design to support model integration into the target application. The work~\cite{lin18} further improved upon the design flow in~\cite{lin17} to devise a customized ensemble modeling method and an integration strategy to boost the accuracy of power prediction. These methods target power model establishment with an offline sampling strategy, which incurs limited adaptability of the created predictors and a long development period. More specifically, the offline power modeling flow is not able to deal with data streams with changing statistical distribution which is known as concept drift~\cite{streamdm}. Moreover, the relatively long development time of the offline power models hinders efficient power model deployment.

In light of these problems, we investigate how our proposed online learning system can be used for FPGA run-time power estimation. With the proposed online learning system, the power models do not need to be completely determined before the applications are implemented onboard, and instead, the applications' power characteristics can be learned during real execution. Furthermore, the power models developed offline can be used as pre-trained models during online power modeling. This section describes the corresponding CAD flow.

\vspace{-4mm}
\subsection{Review of Offline FPGA Power Modeling}
\begin{figure}[t]
	\begin{center}
		\includegraphics[width=\linewidth]{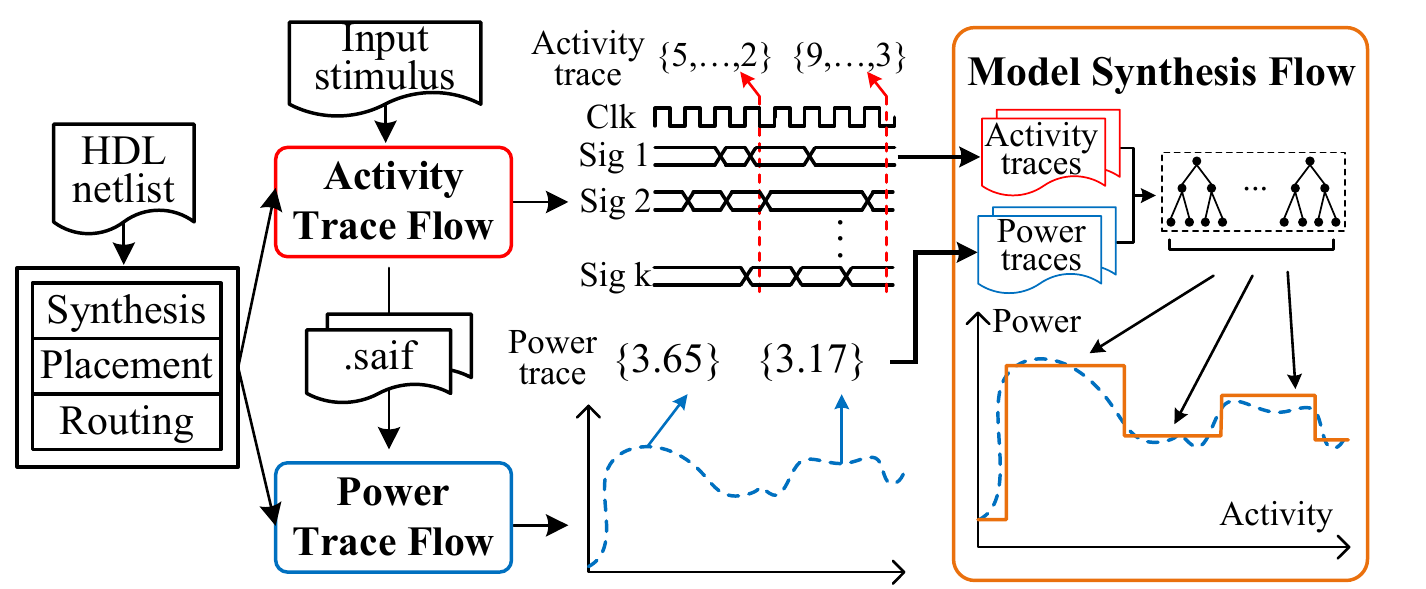}
		\vspace{-4mm}
		\caption{The overall CAD flow for offline FPGA power modeling~\cite{lin18}.}
		\label{fig: dt_pwr}
	\end{center}
\end{figure}

We review the basics of the offline FPGA power modeling flow~\cite{lin18}, as depicted in Fig.~\ref{fig: dt_pwr}. To start with, a given design should pass through synthesis, placement and routing to be transformed into circuit-level representation. Next, the power modeling flow is executed, which comprises three subflows: 1) activity trace flow; 2) power trace flow; and 3) model synthesis flow. In the first flow, a set of signals are identified and monitored to produce attributes as power indicators. The extracted signal activities in a period form an activity trace. At the same time, power simulation ($.saif$) files are generated during simulation, which are used to perform power estimation using vendor tools (e.g., Vivado power analyzer). Following these two flows, the model synthesis flow takes the activity traces and power traces as input, conducts attribute selection, state clustering, hyperparameter tuning, and model training/ensemble. Finally, the offline power models are integrated into the target designs for run-time power prediction.

\vspace{-4mm}
\subsection{Online FPGA Power Modeling}
The CAD flow of FPGA run-time power monitoring with online learning is shown in Fig.~\ref{fig: odtpwr}. This design flow shares the activity trace flow and power trace flow with the offline modeling method~\cite{lin17, lin18}. Herein, the activity traces and power traces are collected to feed in the model generation flow. The model generation flow encompasses attribute selection, power clustering, hardware-aware parameter tuning and model integration, which determines the parameters related to the overall architecture of the online learning system, and then creates and integrates the model into the target design. Note that even though the tree architecture is determined in design time, the model training process has not been conducted at this stage, which differentiates this online power monitoring flow from the offline power modeling flow~\cite{lin17, lin18}. 

\vspace{-3mm}
\subsection{Model Generation Flow}
\label{subsec: pmgf}
The model generation flow consists of four subflows: 1) attribute selection; 2) power clustering; 3) hardware-aware parameter tuning; and 4) model integration. In the remainder of this section, we illustrates each of the above subflows individually.

\textbf{Attribute Selection.} In the activity trace flow, we extract a series of signals with high switching activities to produce attributes as power indicators, based on the heuristic that the signals with higher activities tend to show a richer body of behaviors matching the power patterns. However, we also note that attributes with high switching activities may be correlated (e.g., an input and an output of the same LUT), or exhibit repetitive patterns (e.g, the clock signal). Simply using signals with high activities is not able to guarantee the quality of the extracted attributes. As a result, we identify the attribute quality by adopting an attribute selection method to filter out redundant attributes. Specifically, \emph{recursive attribute elimination} is used. Taking the complete attribute set as the input, the recursive attribute elimination method firstly trains a decision tree model with all attributes, and ranks different attributes by a criterion to quantify attribute importance, such as the Gini impurity in CART decision tree~\cite{cart}. The attributes with least importance are pruned away. The number of attributes is constrained by the hardware-aware parameter tuning.

\begin{figure}[t]
	\begin{center}
		\includegraphics[width=0.96\linewidth]{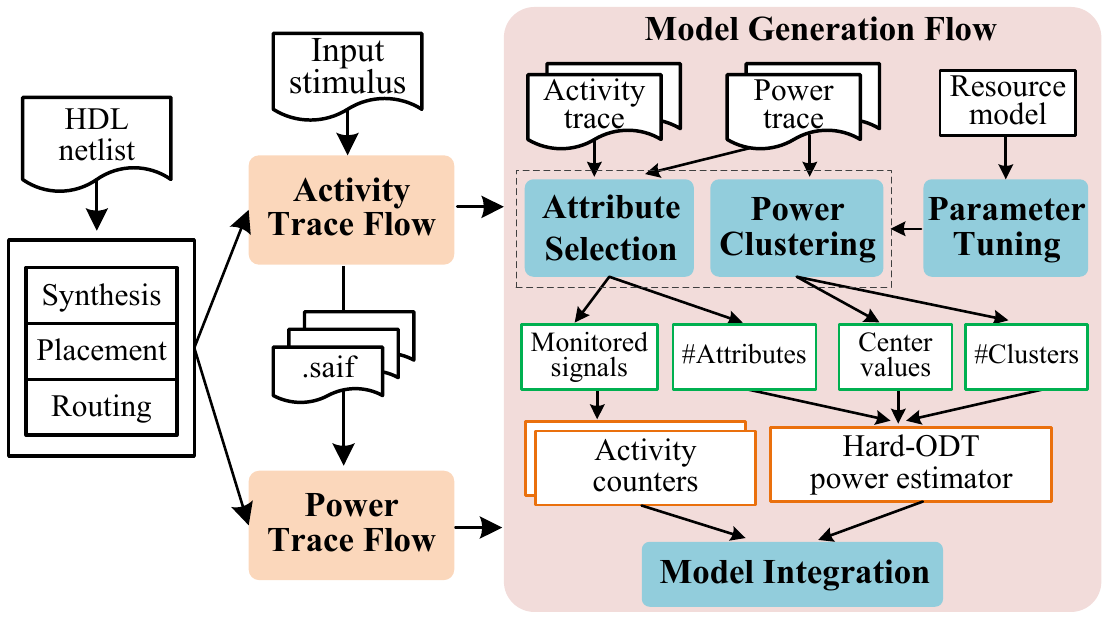}
		\vspace{-3mm}
		\caption{The overall CAD flow for our proposed online power modeling.}
		\label{fig: odtpwr}
	\end{center}
\end{figure}

\textbf{Power Clustering.} We note that the up-to-date power management techniques~\cite{nunez16, losch16} do not require the precise power values for decision making, and therefore, some errors induced in the power monitoring schemes are allowed. Based on this observation, we implement a power clustering stage following the attribute selection in order to trim down the complexity of power representation. This power clustering flow brings two main benefits. Firstly, the resource utilization of the model implementation can be significantly reduced. The complexity of decision tree hardware implementation in terms of classification and regression is different. The decision tree classification employs Gini impurity in CART algorithm~\cite{cart} as shown in Eq.~\ref{eq: ginigain}, information gain in ID3 algorithm~\cite{idt3} or gain ration in C4.5~\cite{quin14} as the split criteria. These split criteria only require the knowledge of the sample distributions. However, for decision tree regression, the split criteria are the standard deviation reduction~\cite{elena12} or decrease in variance~\cite{ikon11}. These split criteria for regression necessitate the computation of mean and variance before and after the split at each split point, and require that each sample value to be recorded for this computation, thus introducing larger resource overhead regarding both memory and computation compared to distribution computation in classification. Our optimized hardware implementation for online decision tree classification algorithm can be applied seamlessly after converting the problem formulation from regression to classification through power clustering. Secondly, by incorporating the power clustering stage, we exert additional control to the resource overhead by parameterizing the number of classes in power monitoring, i.e., the number of clusters for power values. 

\begin{figure}[t]
	\begin{center}
		\includegraphics[width=\linewidth]{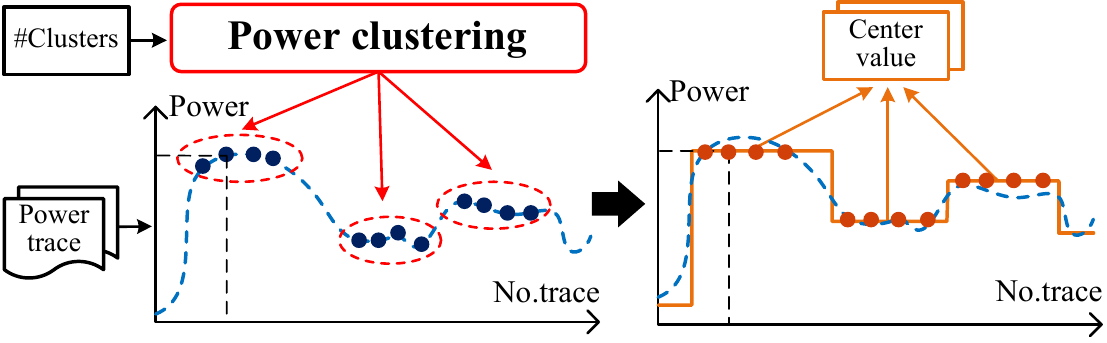}
		\vspace{-6mm}
		\caption{Power clustering flow.}
		\label{fig: pwr_cluster}
	\end{center}
\vspace{-2mm}
\end{figure}

We apply \emph{k-means clustering} on the original power traces from power estimation of FPGA vendor tools, as shown in Fig.~\ref{fig: pwr_cluster}. Then, we replace the original power value in each power trace with the center value of the cluster it belongs to. To determine the number of clusters offering the best performance, we use the Silhouette score~\cite{silho87} as the evaluation metric, while taking into account the constraints set in the following hardware-aware parameter tuning. It also gives an option for the designers to set the number of clusters under different requirements of power granularity/resource usage. 

\textbf{Hardware-Aware Parameter Tuning.} The online decision tree implementation may result in high memory usage as well as DSP usage. To avoid excessive overhead of this additional monitoring hardware, we leverage the models proposed in Section~\ref{sec: odt_rm} to achieve hardware-aware parameter tuning during attribute selection and power clustering. We focus on the optimization of BRAM utilization which is the bottleneck as indicated by the experiments in Section~\ref{sec: exp}. Firstly, the BRAM is widely used in the decision tree inference engine for storing node information in different levels. We observe through experiments that a shallow decision tree is usually enough for power prediction. As a result, to maintain desirable performance while incurring acceptable overheads, we adopt a maximum tree depth of seven for the inference engine design. Secondly, as described in Section~\ref{subsec: bram}, the BRAM utilization is jointly determined by the number of labels ($L$), the number of numeric attributes ($N$), the number of categorical attributes ($C$), the number of quantiles ($Q$), and the number of elements ($E$). We need to keep a balance among all these factors to maintain a small footprint for the generated hardware. To keep the BRAM utilization below 20\% of the design, we set $L \leq 5$, $N \leq 8$, $E = 64$, and $Q = 8$. We constrain the attribute selection and power clustering in Section~\ref{subsec: pmgf} to comply with these requirements, and we fine tune the parameters by evaluating the model accuracy through cross-validation.

\textbf{Model Integration.} At this stage, we have obtained the list of signals to monitor and the number of attributes from the attribute selection stage, and we have also determined the cluster center values and the number of clusters from the power clustering stage. With these parameters defined, the Hard-ODT for power monitoring can then be constructed as depicted in Fig.~\ref{fig: odtpwr}. To capture signal activities, we instrument an activity counter~\cite{lin18} for each of the selected signals in the target design to capture their toggle rates in real time. These activity counters bring negligible effect to the design as reported in~\cite{lin18}. The power monitoring engine, Hard-ODT, together with the activity counters are integrated into the target design to implement and run onboard. 

\vspace{-3mm}
\section{Experiments}
\label{sec: exp}

\begin{table*}[t]
	\centering
	\footnotesize
	\begin{threeparttable}
		\caption{Inference accuracy using different numbers of quantiles.}
		\label{table: acc}
		\begin{tabular}[width=\linewidth]{c|c|c|c|c|c|c|c|c|c|c}
			\toprule
			\multirow{2}{*}{\textbf{Dataset}} & \multicolumn{1}{c|}{\textbf{Gaussian}} & \multicolumn{9}{c}{\textbf{Quantile method with different quantile size}} \\
			& \multicolumn{1}{c|}{\textbf{method}} & \multicolumn{1}{c|}{2} & \multicolumn{1}{c|}{4} & \multicolumn{1}{c|}{8} & \multicolumn{1}{c|}{16} & \multicolumn{1}{c|}{32} & \multicolumn{1}{c|}{64} & \multicolumn{1}{c|}{128} & \multicolumn{1}{c|}{256} & \multicolumn{1}{c}{512}\\
			\midrule
			\multicolumn{1}{c|}{Bank} & 89.10\% & 88.79\% & 89.05\% & 89.15\% & 89.30\% &  89.32\% & 89.26\% & 88.52\% & 88.66\% & 88.59\% \\
			\multicolumn{1}{c|}{Telescope} & 76.16\% & 76.68\% & 74.61\% & 76.41\% & 76.12\% & 76.64\% & 75.51\% & 75.75\% & 76.75\% & 71.32\% \\
			\multicolumn{1}{c|}{Electricity} & 76.26\% & 76.97\% & 77.26\% & 78.02\% & 76.31\% & 77.53\% & 76.91\% & 76.75\% & 76.61\% & 74.15\% \\
			\multicolumn{1}{c|}{Covertype} & 71.02\% & 72.46\% & 72.17\% & 72.72\% & 72.51\% & 71.86\% & 73.43\% & 71.90\% & 70.94\% & 69.41\%\\
			\multicolumn{1}{c|}{Person} & 39.00\% & 45.90\% & 48.82\% & 51.38\% & 52.49\% & 52.35\% & 52.40\% & 47.94\% & 47.44\% & 49.60\%\\
			\bottomrule
		\end{tabular}
	\end{threeparttable}
	\vspace{-7mm}
\end{table*}

\begin{table}[t]
	\centering
	\footnotesize
	\begin{threeparttable}
		\caption{Resource Utilization and Frequency of FPGA Designs	.}
		\label{table: resfreq}
		\begin{tabular}[width=\linewidth]{c|c|c|c|c|c|c}
			\toprule		
			\multicolumn{1}{c|}{\textbf{Dataset}} & \multicolumn{1}{c|}{\textbf{Size}} & \multicolumn{1}{c|}{\textbf{LUT\tnote{1}}} & \multicolumn{1}{c|}{\textbf{BRAM\tnote{2}}} & \multicolumn{1}{c|}{\textbf{URAM\tnote{3}}} & \multicolumn{1}{c|}{\textbf{DSP\tnote{4}}} & \multicolumn{1}{c}{\textbf{Freq\tnote{5}}}\\
			\midrule
			\multicolumn{1}{c|}{Bank} & 45211 & 63079 & 486 & 0 & 202 & 308 \\
			\multicolumn{1}{c|}{Telescope} & 19020 & 73800 & 480 & 0 & 184 & 305 \\
			\multicolumn{1}{c|}{Electricity} & 45312 & 54198 & 384 & 0 & 138 & 300 \\
			
			\multicolumn{1}{c|}{Covertype} & 581012 & 169334 & 1822 & 61 & 1126 & 170 \\
			\multicolumn{1}{c|}{Person} &164860 &  59401 & 986 & 0 & 191 & 266 \\
			\bottomrule
		\end{tabular}
		\begin{tablenotes}
			\scriptsize
			\item[1]Total No. LUT: 1182240\item[2]Total No. BRAM36: 2160\item[3]Total No. URAM: 960\item[4]Total No. DSP: 6840\item[5] Measured in the units of MHz
		\end{tablenotes}
	\end{threeparttable}
\end{table}

\subsection{Experimental Setup}
In the experiments, we put our main focus on online tree learning. The differences in traditional, batch and online tree learning have been studied in prior works~\cite{hang10, pedro00} and are not elaborated in this paper. We first implement the software version of our proposed algorithm in StreamDM-C++\cite{streamdm}, the state-of-the-art software toolkit supporting the Hoeffding tree. The parameter settings related to the Hoeffding bound are $n_{min}$ = 200, $n_{pt}$ = 10, $\tau$ = 0.05, $\delta$ = $10^{-3}$ and $\lambda$ = 0.01, according to~\cite{pedro00, streamdm} and~\cite{qt17}. The maximum leaf number is 1024, and the maximum tree depth is 15. We use a 32-bit fixed-point data representation with a 30-bit fraction for numeric attributes, after normalizing the data to within the range of [-1,1], if necessary. We evaluate the design with five large datasets: Bank Marketing (Bank), MAGIC Gamma Telescope (Telescope), Australian New South Wales Electricity Market (Electricity), Covertype and Person Activity (Person) from the UCI machine learning repository~\cite{uci07} and related works~\cite{streamdm, cheng15}. The optimized hardware is designed in Verilog and implemented on the Xilinx VCU1525 platform~\cite{virtexp} using SDAccel 2018.2. Table~\ref{table: resfreq} shows the size of datasets and information about FPGA implementation. The datasets are transferred from CPU to off-chip memory (DDR4) on the FPGA platform through PCIe.

\vspace{-4mm}
\subsection{Tuning the Number of Quantiles}
\label{subsec: qtnum}
We tune the number of quantiles in a wide range to evaluate the model performance. The evaluation methodology is \emph{Interleaved-test-then-train}: each sample is first passed through testing before it is applied for training. This is a commonly used evaluation method for online learning models, and the model performance is evaluated by inference accuracy for the entire datasets. In this way, both the online training and testing phases fully utilize the whole datasets, which is different from offline training methods that require a train-test division and need to separately evaluate training and testing accuracy. 

Experimental results in Table~\ref{table: acc} show that the inference accuracy may be degraded significantly as the number of quantiles becomes either too small or too large, especially for the Person dataset. When the quantile number is small, the learning ability of the model may be constrained, because the learned distribution is too coarse-grained to provide effective information. Conversely, if the quantile number becomes too large, the generalization ability may be impaired as well, since the design is more prone to noise in the datasets. Setting the quantile number between 8 and 32 provides high accuracy with desirable robustness. Considering the fact that memory and computation demand is proportional to the number of quantiles, we adopt a unified quantile number of 8 in the hardware design. One can also tune the quantile number to best fit a target dataset.

\vspace{-5mm}
\subsection{Comparison with Batch Learning on FPGA}
The up-to-date method to cope with decision tree learning with large datasets on FPGA is through batch learning. The work~\cite{cheng15} presented a state-of-the-art FPGA architecture for batch-based decision trees. Covertype is used as the only benchmark in~\cite{cheng15}, and it serves as the baseline for comparison in Table~\ref{table: compbatch}. The accuracy and overall resource usage are not given, but study in~\cite{pedro00} has proven that both Hoeffding tree and batch tree can lead to the same results for large datasets asymptotically. Table~\ref{table: compbatch} shows that our proposed online learning design can offer an up to 4-orders-of-magnitude speedup in execution time in comparison to~\cite{cheng15}. This significant speedup stems from the difference in communication patterns. The work~\cite{cheng15} involves a number of rounds of transmission for the same samples from and to the off-chip DDR memory in the training process per batch: it reads the sample set at the start of a split process and writes back the subset of samples in each resulting split. By contrast, our proposed online training architecture only requires reading each sample once in the entire learning process, thus reducing a large amount of high-cost inter-chip communication.

\begin{table}[t]
	\centering
	\footnotesize
	\caption{Performance comparison: batch learning \& online learning.}
	\label{table: compbatch}
	\vspace{-2mm}
	\begin{tabular}[width=\linewidth]{c|c|c|c}
		\toprule
		\multicolumn{1}{c|}{\textbf{Method}} & \multicolumn{1}{c|}{\textbf{Platform}} & \multicolumn{1}{c|}{\textbf{Freq.}} & \multicolumn{1}{c}{\textbf{Exe. time}}\\
		\midrule
		\multicolumn{1}{c|}{Batch learning~\cite{cheng15}} & Intel Stratix IV  & 200 MHz & 118 s \\
		\multicolumn{1}{c|}{This work} & Xilinx Ultrascale+ & 170 MHz & 3.97 ms \\
		\bottomrule
	\end{tabular}
\end{table} 

\begin{figure*}[t]
	\begin{center}
		\includegraphics[width=5.15cm]{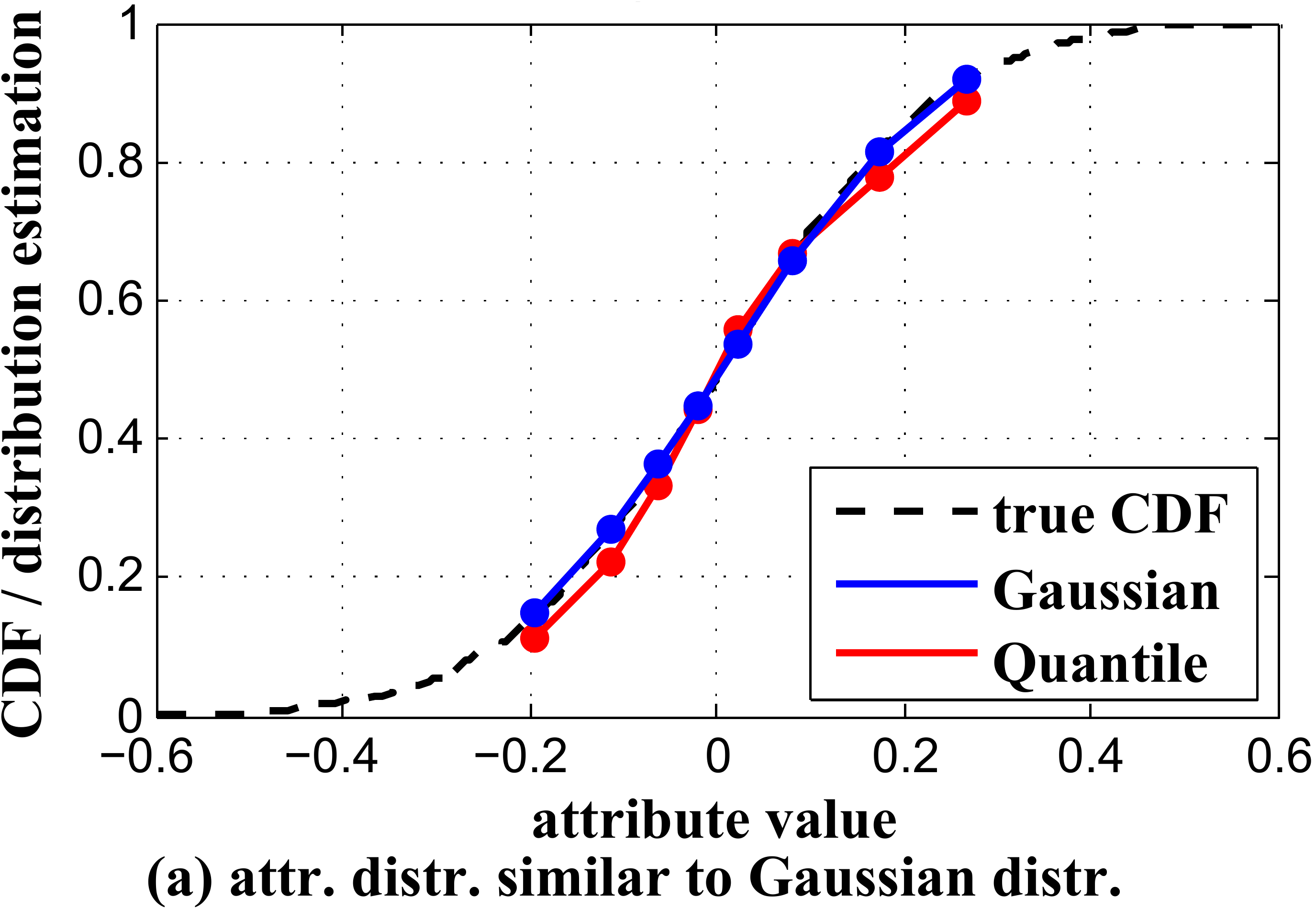}
		\includegraphics[width=5.15cm]{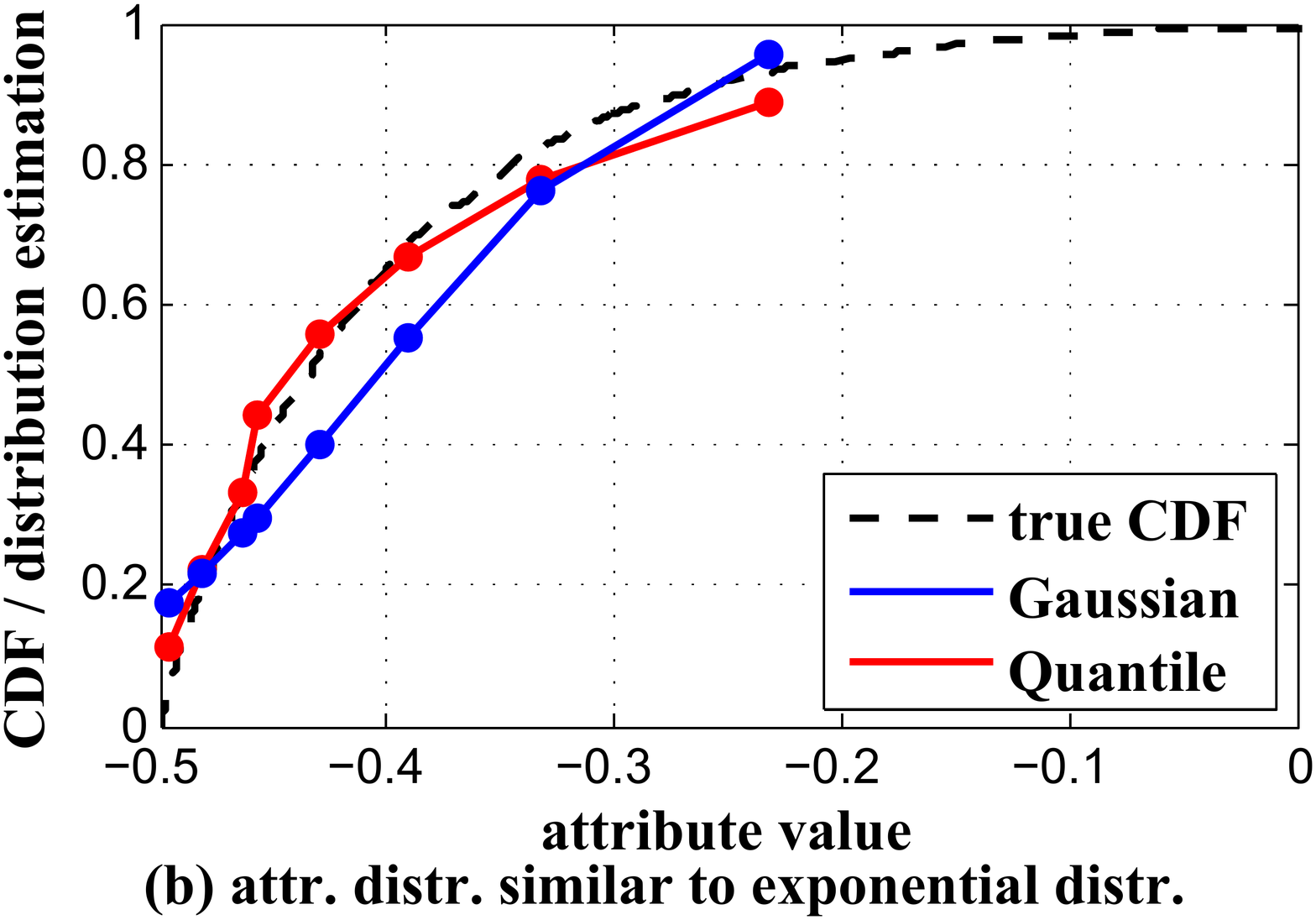}
		\includegraphics[width=5.15cm]{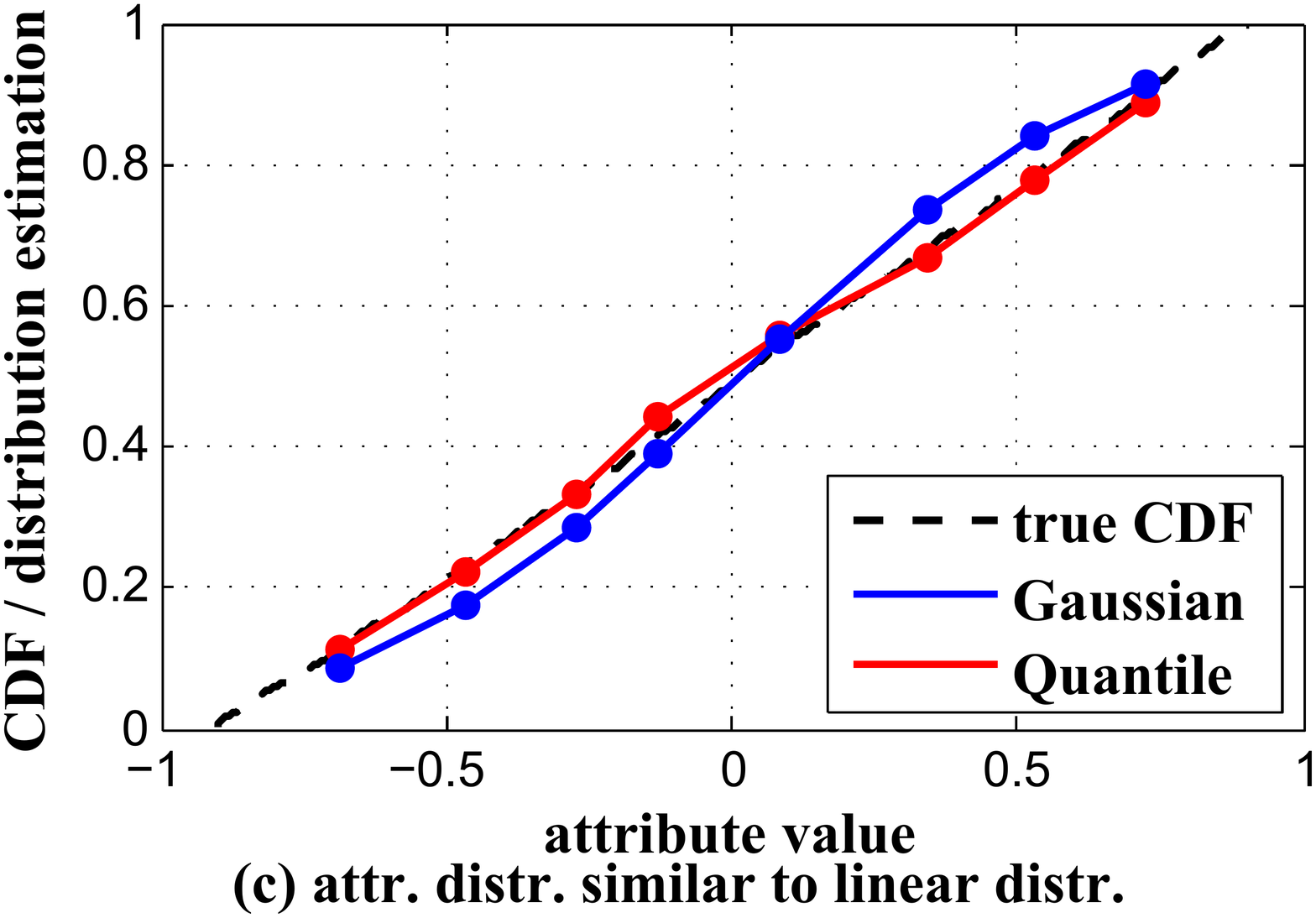}
		\vspace{-3mm}
		\caption{Gaussian and quantile estimation of true CDFs of three representative attributes with different statistical distributions from Electricity dataset.}
		\label{fig: apprx}
	\end{center}
	\vspace{-8mm}
\end{figure*}

\vspace{-4mm}
\subsection{Comparison with the State-of-the-art on Processors}
\label{subsec: vscpu}
StreamDM-C++~\cite{streamdm} reported that Gaussian method provided the best performance amongst prior methods~\cite{vfml, vfdtc, ga08, streamdm}, so it is used as the baseline in this paper. Regarding inference accuracy, our proposed algorithm with eight quantiles outperforms the Gaussian method for all five benchmarks, with 0.05\% to 12.3\% improvement, as shown in Table~\ref{table: acc}. 

The results of CDF approximation using the quantile method and Gaussian method account for this gap in accuracy. Three attributes with representative statistical distributions in the Electricity dataset are selected to illustrate the results, as shown in Fig.~\ref{fig: apprx}. The sample set is the subset in the root node before it is split. The CDF of the first attribute is close to the Gaussian function, and thereby, the Gaussian method provides slightly better fitting results than the quantile method. However, regarding the second and third attributes, the quantile method outperforms the Gaussian method. The Gaussian method assumes that the sample distribution conforms with Gaussian distribution, and lead to poor approximation quality for distributions dissimilar to Gaussian. By contrast, the quantile method makes no presumption of any distribution, and hence, it offers accurate approximation for various distributions. In other words, the quantile method has a wider scope of applicability than the Gaussian method, which accounts for the improvement in accuracy.

For the execution time, we integrate the quantile method in StreamDM-C++ and run this toolkit with both the Gaussian and quantile methods on the Xeon E5-2680 platform under 2.6 GHz. As shown in Table~\ref{table: exetime}, our proposed hardware designs on FPGA achieve 423$\times$ to 1526$\times$ speedup over the Gaussian method and 384$\times$ to 1581$\times$ speedup over the quantile method in software implementation, respectively.

\begin{table}[t]
	\centering
	\footnotesize
	\caption{Comparison of Software and Hardware Execution Time.}
	\label{table: exetime}
	\vspace{-2mm}
	\begin{tabular}[width=\linewidth]{c|c|c|c|c}
		\toprule
		\multirow{2}{*}{\textbf{Dataset}} & \multicolumn{2}{c|}{\textbf{CPU exe. time}} & \multicolumn{1}{c|}{\textbf{FPGA}}  
		& \multirow{2}{*}{\textbf{Speedup}}\\
		& \multicolumn{1}{c|}{Gaussian} & \multicolumn{1}{c|}{Quantile} & \multicolumn{1}{c|}{\textbf{exe. time}} 
		& \multicolumn{1}{c}{}\\
		\midrule
		\multicolumn{1}{c|}{Bank} & 0.27 s & 0.25 s & 0.36 ms & 750 / 694 $\times$ \\
		\multicolumn{1}{c|}{Telescope} & 0.11 s & 0.10 s & 0.26 ms & 423 / 384 $\times$ \\
		\multicolumn{1}{c|}{Electricity} & 0.21 s & 0.20 s & 0.42 ms & 500 / 476 $\times$ \\
		\multicolumn{1}{c|}{Covertype} & 6.06 s & 6.28 s & 3.97 ms & 1526 / 1581 $\times$ \\
		\multicolumn{1}{c|}{Person} & 0.79 s & 0.75 s & 0.93 ms & 849 / 806 $\times$ \\
		\bottomrule
	\end{tabular}
\end{table}

\vspace{-4mm}
\subsection{Performance and Resource Modeling}
We evaluate the accuracy of performance and resource models proposed in Section~\ref{sec: odt_pm} and~\ref{sec: odt_rm}, respectively. Results are shown in Table~\ref{table: acc_model}, and the corresponding real values of performance and resource metrics are described in Table~\ref{table: exetime} and Table~\ref{table: resfreq}, respectively. Experimental results demonstrate the correctness of design profiling and the efficacy of our performance and resource models in evaluating the execution time, DSP and BRAM utilization, with average modeling accuracy reaching up to 94.62\%, 100\% and 98.78\%, respectively. These analytical models offer early and fast performance/resource evaluation of the resulted hardware designs, which can significantly expedite the process of trading off between different design metrics and selecting the suitable devices for implementation.

\begin{table}[t]
	\centering
	\footnotesize
	\caption{Accuracy of Execution Time, DSP and BRAM Modeling.}
	\label{table: acc_model}
	\vspace{-1mm}
	\begin{tabular}[width=\linewidth]{c|c|c|c}
		\toprule
		\multirow{2}{*}{\textbf{Dataset}} & \multicolumn{1}{c|}{\textbf{Performance}} & \multicolumn{2}{c}{\textbf{Resource utilization}}\\
		& \multicolumn{1}{c|}{Exe. time} & \multicolumn{1}{c|}{DSP} & \multicolumn{1}{c}{BRAM}\\
		\midrule
		\multicolumn{1}{c|}{Bank} & 95.43\% (0.37ms) & 100\% (202) & 99.28\% (482.5) \\
		\multicolumn{1}{c|}{Telescope} & 88.35\% (0.29ms) & 100\% (184) & 99.06\% (475.5) \\
		\multicolumn{1}{c|}{Electricity} & 92.11\% (0.39ms) & 100\% (138) & 97.92\% (376) \\
		\multicolumn{1}{c|}{Covertype} & 99.73\% (3.98ms) & 100\% (1126) & 98.70\% (1907.5) \\
		\multicolumn{1}{c|}{Person} & 97.50\% (0.91ms) & 100\% (191) & 98.94\% (975.5) \\
		\midrule
		\multicolumn{1}{c|}{Average} & 94.62\% & 100\% & 98.78\% \\
		\bottomrule
	\end{tabular}
\end{table}

\vspace{-4mm}
\subsection{FPGA Power Monitoring with Online Learning}
We collect 40000 samples to evaluate the accuracy and resource overhead of our online modeling method for run-time power on FPGA. To determine the architecture parameters of the online decision tree models, only the first 5000 samples are used, while the others are used to train the model in real time. As a comparison, we build offline decision tree models~\cite{lin18}, with 32000 samples used for model parameter tuning. To compare these two cases fairly, 80\% of the samples are used to train the models and the rest 20\% are used for testing. We set the sampling period to be 3 $\mu$s. The benchmarks we used are from different application categories of Polybench~\cite{polybench}. 

The power modeling accuracy and the corresponding resource overheads regarding online and offline models are shown in Table~\ref{table: resaccodt}. In Table~\ref{table: resaccodt}, the accuracy for online decision tree is on par with that of the traditional decision tree, with only a 0.03\% difference. The results verify that the online decision tree approaches the traditional decision tree when the sample size is large enough~\cite{hang10, pedro00}. Regarding resource utilization, the online decision tree model consumes on average 3.63\% of LUT, 2.40\% of FF, 13.39\% of BRAM and 2.04\% of DSP. It is worth noting that the online decision tree model requires only 15.6\% of samples for offline modeling since these samples are only used to select suitable attributes and determine the tree architectures. This speedup in development time is accompanied by moderately larger resource overheads. Moreover, the proposed online decision tree learning method can make use of the offline power models as pre-trained models and it is able to learn from samples with various statistical distributions, as demonstrated in Section~\ref{subsec: vscpu}, which distinguishes itself from the offline modeling approach with higher efficacy.


\begin{table}[t]
	\centering
	\footnotesize
	\begin{threeparttable}
		\caption{Resource Overheads and Accuracy Using Hard-ODT for FPGA run-time power monitoring.}
		\label{table: resaccodt}
		\begin{tabular}[width=\linewidth]{c|c|c|c|c|c|c}
			\toprule
			\multirow{2}{*}{\textbf{Dataset}} & \multicolumn{4}{c|}{\textbf{Resource utilization}} & \multicolumn{2}{c}{\textbf{Accuracy}} \\
			& \multicolumn{1}{c|}{LUT} & \multicolumn{1}{c|}{FF} & \multicolumn{1}{c|}{BRAM} & \multicolumn{1}{c|}{DSP} & \multicolumn{1}{c|}{Offline} & \multicolumn{1}{c}{Online} \\
			\midrule
			\multicolumn{1}{c|}{Atax} & 50971 & 67663 & 349   & 166 & 94.1\%  & 94.1\%  \\
			\multicolumn{1}{c|}{Gemm} & 42616 & 56043 & 286   & 139 & 93.42\% & 93.38\% \\
			\multicolumn{1}{c|}{Symm} & 23522 & 31098 & 155.5 & 76  & 97.98\% & 97.96\% \\
			\multicolumn{1}{c|}{Syrk} & 38538 & 51038 & 250.5 & 124 & 97.98\% & 97.96\% \\
			\multicolumn{1}{c|}{Mvt}  & 59223 & 77476 & 405   & 193 & 92.03\% & 91.97\% \\
			\midrule
			\multicolumn{1}{c|}{Average} & 42974 & 56664 & 289 & 140 & 95.10\% & 95.07\% \\
			\bottomrule
		\end{tabular}
	\end{threeparttable}
\end{table}

\section{Conclusion}
Online decision tree algorithms suffer from either high memory usage or high computational intensity with dependency and long latency. In this paper, we introduce an efficient and scalable quantile-based induction algorithm for the Hoeffding tree, and we investigate hardware optimization techniques specific to this algorithm. After that, we build Hard-ODT, a hardware-friendly online decision tree learning system with system-level optimizations. Furthermore, a performance model and a resource model are proposed for early evaluation of design metrics and trade-off between performance and resource. 
Finally, we investigate how the proposed online learning system can be used for FPGA run-time power monitoring as a case study. Experimental results show that our design remarkably reduces memory and computational demand, showing 384$\times$ -- 1581$\times$ speedup in execution time over the state-of-the-art design while achieving 0.05\% -- 12.3\% improvement in accuracy, which enables the online decision trees to be used for applications requiring fast response time, and makes it more efficient for online decision tree architecture search. Regarding power modeling efficacy, the proposed online power modeling strategy is on par with the traditional offline power modeling method, whereas it requires a much smaller number of samples to be collected. Moreover, the quantile-based algorithm-hardware co-design methodology can also benefit a wide range of machine learning methods, such as ensemble learning, quantile regression and imbalanced dataset resampling.

\vspace{-3mm}
\bibliographystyle{IEEEtran}
\bibliography{ref}

\vspace{-8mm}

\begin{IEEEbiography}[{\includegraphics[width=1in,height=1.25in,clip,keepaspectratio]{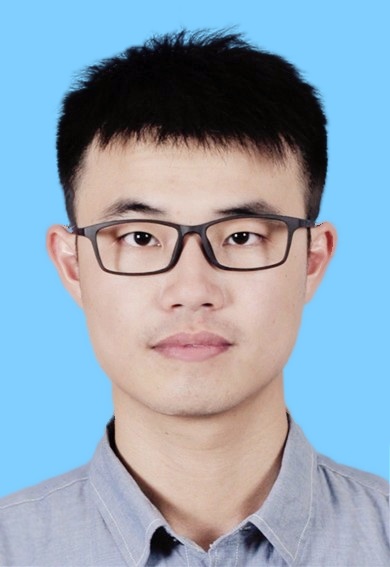}}]{Zhe Lin}
(S'15, M'20) received his B.S. degree from School of Electronic Science and Engineering from Southeast University, China (2014). He then received the Ph.D. degree from the Department of Electronic and Computer Engineering at Hong Kong University of Science and Technology, Hong Kong (2019). From 2020, he has been a Research Associate in Peng Cheng Laboratory, China. Zhe's research interests cover FPGA power prediction and optimization, and hardware-aware AI implementation.
\end{IEEEbiography}

\vspace{-12mm}

\begin{IEEEbiography}[{\includegraphics[width=1in,height=1.25in,clip,keepaspectratio]{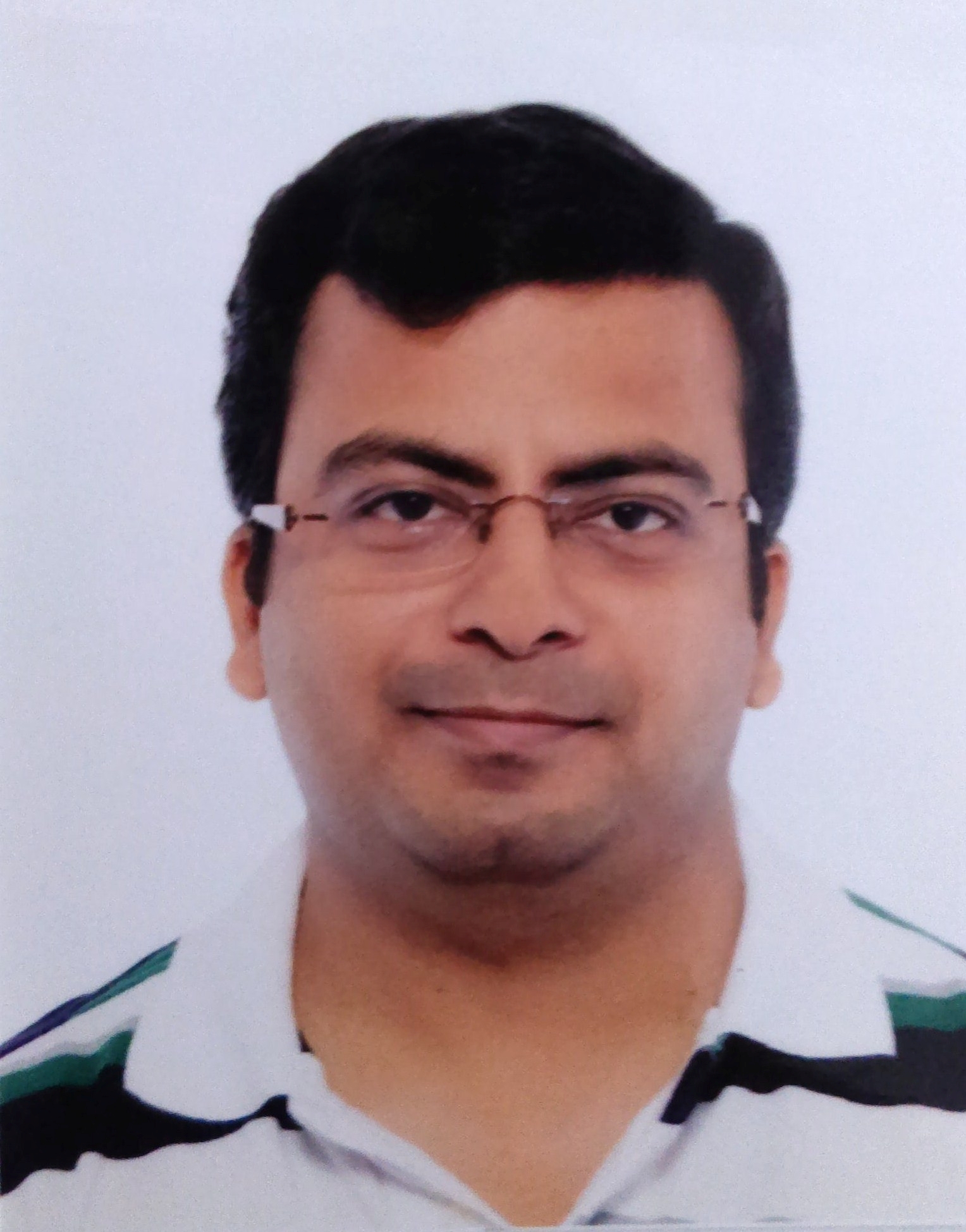}}]{Sharad Sinha}
(S'03, M'14) is an assistant professor with Dept. of Computer Science and Engineering, Indian Institute of Technology (IIT) Goa. Previously, he was a Research Scientist at NTU, Singapore. He received his PhD degree in Computer Engineering from NTU, Singapore (2014). He received the \textit{Best Speaker Award} from \textit{IEEE CASS Society}, Singapore Chapter, in 2013 for his PhD work on High Level Synthesis and serves as an Associate Editor for \textit{IEEE Potentials} and \textit{ACM Ubiquity}. Dr. Sinha earned a Bachelor of Technology (B.Tech) degree in Electronics and Communication Engineering from Cochin University of Science and Technology (CUSAT), India in 2007. From 2007-2009, he was a design engineer with Processor Systems (India) Pvt. Ltd. Dr. Sinha's research and teaching interests are in computer arhcitecture, embedded systems and reconfigurable computing.
\end{IEEEbiography}

\vspace{-12mm}

\begin{IEEEbiography}[{\includegraphics[width=1in,height=1.25in,clip,keepaspectratio]{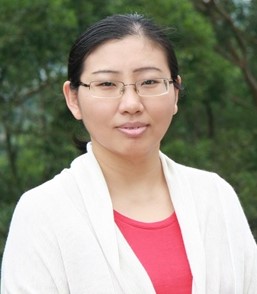}}]{Wei Zhang}
(M'05) received a Ph.D. degree from Princeton University, Princeton, NJ, USA, in 2009. She was an assistant professor with the School of Computer Engineering, Nanyang Technological University, Singapore, from 2010 to 2013. Dr. Zhang joined the Hong Kong University of Science and Technology, Hong Kong, in 2013, where she is currently an associated professor and she established the reconfigurable computing system laboratory (RCSL).

Dr. Zhang has authored or co-authored over 80 book chapters and papers in peer reviewed journals and international conferences. Dr. Zhang serves as the Associate Editor for \textit{TECS}, \textit{TVLSI}, and \textit{JETC}. She also serves on many organization committees and technical program committees. Dr. Zhang’s current research interests include reconfigurable systems, FPGA-based design, low-power high-performance multicore systems, electronic design automation, embedded systems, and emerging technologies.
\end{IEEEbiography}

\end{document}